\documentclass[11pt]{article}

\usepackage[final]{acl}
\usepackage{graphicx}
\usepackage{subcaption}
\usepackage{wrapfig}
\usepackage{amsmath}
\usepackage{booktabs}   
\usepackage{multirow}   
\usepackage{graphicx}   
\usepackage[table]{xcolor} 
\usepackage{hyperref}
\usepackage{url}
\usepackage[linesnumbered,ruled,vlined]{algorithm2e}
\usepackage[utf8]{inputenc} 
\usepackage[T1]{fontenc} 
\usepackage{hyperref}   
\usepackage{url}      
\usepackage{booktabs} 
\usepackage{tabularx}
\usepackage{amsfonts}  
\usepackage{nicefrac} 
\usepackage{microtype} 
\usepackage{xcolor}    
\usepackage{graphicx}
\usepackage{multicol}
\usepackage{array}
\usepackage{svg}
\usepackage{multirow}
\usepackage{enumitem}
\usepackage{ dsfont }
\usepackage{amsthm}
\usepackage{adjustbox}
\usepackage{ amssymb }
\usepackage{bm}

\usepackage{times}
\usepackage{latexsym}

\usepackage[T1]{fontenc}

\usepackage[utf8]{inputenc}

\usepackage{microtype}

\usepackage{inconsolata}

\usepackage{graphicx}

\title{AIM-CoT: Active Information-driven Multimodal Chain-of-Thought for Vision-Language Reasoning}

\author{
 \textbf{Xiping Li\textsuperscript{1}},
 \textbf{Jianghong Ma\textsuperscript{2}}\thanks{Corresponding Author}
\\
\\
 \textsuperscript{1}The Chinese University of Hong Kong \\
 \textsuperscript{2}Harbin Institute of Technology (Shenzhen)
\\
 \textit{{lihsiping@gmail.com}, {majianghong@hit.edu.cn}
 }
}

\begin{document}
\maketitle

\begin{abstract}
Interleaved-Modal Chain-of-Thought (I-MCoT) advances vision-language reasoning, such as Visual Question Answering (VQA). This paradigm integrates specially selected visual evidence from the input image into the context of Vision-Language Models (VLMs), enabling them to ground their reasoning logic in these details.
Accordingly, the efficacy of an I-MCoT framework relies on identifying \textit{what} to see (evidence selection) and \textit{when} to see it (triggering of insertions).
However, existing methods fall short in both aspects. 
First, for selection, they rely on attention signals, which are unreliable---particularly under severe granularity imbalance between the brief textual query and the informative image.
Second, for triggering, they adopt static triggers, which fail to capture the VLMs' dynamic needs for visual evidence.
To this end, we propose 
a novel I-MCoT framework, \textbf{A}ctive \textbf{I}nformation-driven \textbf{M}ulti-modal \textbf{C}hain-\textbf{o}f-\textbf{T}hought (\textbf{AIM-CoT}),
which aims to improve both evidence selection and insertion triggering via: (1) \textbf{Context-enhanced Attention-map Generation (CAG)} to mitigate granularity imbalance via textual context enhancement;
(2) \textbf{Active Visual Probing (AVP)} to proactively select the most informative evidence via an information foraging process;
and 
(3) \textbf{Dynamic Attention-shift Trigger (DAT)} to precisely activate insertions when VLM's attention shifts from text to visual context. 
Experiments across three benchmarks and four backbones demonstrate AIM-CoT's consistent superiority. Our code is available at \url{https://anonymous.4open.science/r/AIMCoT}.
\end{abstract}

\section{Introduction}\label{sec: introduction}

Chain-of-Thought (CoT) prompting, initially established for Large Language Models (LLMs) \cite{cot1,cot2,cot3,cot4,cot5,cot6,cot7,cot9,weizhu}, has been naturally adapted to Vision-Language Models (VLMs) \cite{cot_vlm_1,cot_vlm_2,cot_vlm_3,cot_vlm_4}, empowering them to tackle vision-related tasks via a series of intermediate reasoning steps.
{Visual Question Answering (VQA)} serves as a representative scenario targeted in this study. In a typical VQA scenario, the VLM is presented with a multimodal question comprising a {textual query} and an associated {image}, and is then prompted to answer the question.

Early efforts in this domain \cite{ccdot, ddcot, scaffold, mmcot} focused on generating text-only CoT. A pivotal advancement is the paradigm of \textbf{Interleaved-Modal Chain-of-Thought (I-MCoT)} \cite{icot}. Unlike text-only methods, this paradigm aims to provide fine-grained visual evidence for model reasoning. Specifically, it first selects the salient sub-regions from the input image, and then inserts them as visual tokens into the context of reasoning chain.

However, to construct an effective I-MCoT framework, two critical questions should be well addressed:  \textbf{(1) What to see?} (Selection: identifying the specific image regions that support the current reasoning step) and \textbf{(2) When to see it?} (Triggering: determining the precise moment visual insertion is needed).

Existing research \cite{icot} falls short in these two aspects due to its reliance on passive and static heuristics. Regarding \textbf{selection}, it adopts attention-based selection, which relies on the quality of cross-attention maps of VLM layers. Specifically, the visual regions that receive the most attention from tokens in the text context are selected, assumed to be most helpful for the subsequent generation. However, as revealed in Section \ref{sec: motivation for idea 1}, the raw VLM attention fails to accurately identify the salient regions, particularly when there is a granularity imbalance. Specifically, in the input context, the rich details in the image overwhelm the brief textual query. Therefore, with limited semantic anchors, the query cannot effectively steer the cross-attention towards the crucial visual regions.
Regarding \textbf{triggering}, prior works often insert visual information at fixed, predefined moments (e.g., upon generating a newline character \cite{icot}). As analyzed in Section~\ref{sec: motivation for idea 3}, such a static mechanism fails to align with the model's dynamic cognitive need for fine-grained visual information.

To overcome these limitations, we propose 
a novel I-MCoT framework, \textbf{A}ctive \textbf{I}nformation-driven \textbf{M}ulti-modal \textbf{C}hain-\textbf{o}f-\textbf{T}hought (\textbf{AIM-CoT}),
which aims to address both \textit{what to see} and \textit{when to see it} by shifting VLM reasoning from passive, static perception to active, dynamic exploration. Grounded in Information Foraging Theory \cite{motivation2_1, motivation2_2}, AIM-CoT detects the VLM's momentary need for visual cues and proactively forages for the most informative evidence. Integrated into the context window, this evidence improves subsequent generation.
AIM-CoT realizes this shift through three synergistic components:
   (1) \textbf{Context-enhanced Attention-map Generation (CAG):} 
   {To refine the VLM's cross-modal attention distribution, CAG elicits a query-conditioned description of the image and appends it to the textual context. Rather than serving as an auxiliary caption, CAG provides semantic anchors that align textual and visual granularities.}
   (2) \textbf{Active Visual Probing (AVP):} To address the {``What to see''} problem, AVP quantifies the information gain of candidate regions, integrating those that provide the highest utility into the context for subsequent reasoning and generation.
    (3) \textbf{Dynamic Attention-shift Trigger (DAT):} To address the {``When to see it''} problem, DAT dynamically and precisely triggers visual insertion when a significant shift from textual to visual focus is detected, indicating the VLM's cognitive demand for visual evidence.

Our contributions are summarized as follows:
\begin{itemize}
[topsep=0.5mm, partopsep=0pt, itemsep=0pt, leftmargin=10pt]  
    \item We introduce AIM-CoT, a unified system (CAG, AVP, DAT) that enables VLMs to proactively forage for informative visual evidence and dynamically capture the critical triggering moments.
    \item Our analyses demonstrate that AIM-CoT (1) selects truly salient visual evidence via information gain, (2) withstands noisy attention and descriptions, and (3) remains deployment-friendly.
    \item Experiments conducted on four backbones across three VQA benchmarks demonstrate the consistently superior performance of AIM-CoT over state-of-the-art baselines.
\end{itemize}

\section{Related Work}\label{sec: related work}

Multimodal CoT has been widely adopted in VLM research. Early text-only efforts enhance reasoning by decomposing questions \cite{ddcot}, generating intermediate scene graphs \cite{ccdot}, or overlaying coordinate grids for spatial referencing \cite{scaffold}.

A pivotal advancement is the I-MCoT paradigm, which integrates visual evidence directly into reasoning chains to improve subsequent reasoning/generation. The leading approach, ICoT \cite{icot}, selects 
the visual patches receiving the highest attention scores from the text tokens in the context. Then, this visual evidence is integrated when the VLM outputs a newline token. 

However, our analysis in Section \ref{sec: motivations} highlights the limitations of existing research. Therefore, while we adopt the established I-MCoT paradigm (similar to ICoT \cite{icot}), our contributions lie in shifting from the passive selection and static trigger to the proactive information-foraging selection and dynamic attention-shifting trigger.

\section{Motivation}\label{sec: motivations}

We focus on two pivotal questions for I-MCoT: \textbf{What} and \textbf{When} to see. First, we expose the limitations of passive attention-based selection in Section~\ref{sec: motivation for idea 1}. To bridge this, we explore an active paradigm shift, investigating the potential of information gain for selection (Section~\ref{sec: motivation for idea 2}) and dynamic attention shifts for triggering (Section~\ref{sec: motivation for idea 3}).

\subsection{{Moving Beyond Attention Maps: The Reliability Gap in Passive Visual Selection}}
\label{sec: motivation for idea 1}

Existing I-MCoT methods adopt attention-based visual selection. We examine the reliability of attention \textbf{for visual selection} by asking:
\begin{enumerate}[label=(\arabic*), topsep=0.5mm, partopsep=0pt, itemsep=0pt, leftmargin=17pt]
\item \textbf{Sufficiency:} Do highly attended patches always benefit question answering?
\item \textbf{Necessity:} Do truly crucial patches always receive high attention?
\end{enumerate}

To investigate these two questions, we randomly sample 500 instances from M3CoT and 500 from ScienceQA, respectively. 

Regarding sufficiency, we mask the top-$K_{mask}$ most attended regions and evaluate the performance drop. Notably, the masking is only for evaluating the importance of these regions as visual evidence, not for AIM-CoT itself. 
As shown in Table~\ref{tab: performance degradation when masked} (Appendix~\ref{app:supplementary for sufficiency check in motivation 1}), masking highly attended regions leads to only minor degradation. Although information redundancy may explain this, our upcoming necessity check suggests that attention peaks rarely coincide with truly salient regions, failing to be sufficient signals for evidence selection.

Regarding necessity, we examine how well the most attended regions $R_{attn}$ cover the truly crucial regions $R_{true}$ in Section~\ref{app:semantic_relevance}, which is quantified by Intersection over Union (IoU). Across 1,000 instances, over 75\% have an IoU even below 50\%. This shows that attention peaks rarely align with truly critical visual evidence, i.e., crucial regions do not necessarily emerge as the most attended ones. Moreover, Appendix~\ref{app:supplementary for granularity imbalance in necessity check in motivation 1} suggests that alleviating the text-vision granularity imbalance is an effective way to improve this alignment.

In conclusion, attention alone is unreliable for evidence selection. Therefore, we do not treat attention as a selection rule.
However, improving \textbf{attention distribution} remains crucial, as it (1) underlies autoregressive generation in VLMs and (2) provides candidate regions for subsequent information-driven selection.
Accordingly, we introduce CAG, which assists VLMs to attend to critical visual content precisely by
enriching brief textual queries with semantic anchors (Section \ref{sec: method 1}).

\subsection{{Identifying What to See: From Semantic Alignment to Information Gain}}
\label{sec: motivation for idea 2}

The empirical limitations observed in Section~\ref{sec: motivation for idea 1} can be attributed to a fundamental misalignment: cross-attention maps primarily capture the \textit{semantic correlations} among tokens, whereas the ultimate goal of the selection phase in I-MCoT is to identify visual evidence that provides \textit{rich visual information} for subsequent reasoning and generation.

This insight motivates a shift from attention-based selection to an information-centric approach. However, two fundamental questions arise regarding the design of such a mechanism: (1) \textbf{Metric}: What constitutes a theoretical measure of ``information'' in the context of VLM reasoning? (2) \textbf{Process}: How should the selection mechanism mimic the cognitive process of information gathering, as opposed to static Top-K selection?

To answer these, we adopt \textbf{Information Foraging Theory (IFT)} \cite{motivation2_1, motivation2_2, motivation2_3, motivation2_4,} as a {guiding principle for our framework design}.
First, regarding the \textit{metric}, IFT suggests that information is valuable only when it reduces the agent's uncertainty~\cite{xfzhou}. This motivates us to ground our selection criterion in uncertainty reduction (i.e., information gain), which can be quantified by the entropy of the VLM's predictive distribution over its vocabulary.
Second, regarding the \textit{process}, IFT characterizes foraging as a sequential trajectory where each step depends on previous knowledge. This suggests that the optimal strategy is not a one-off Top-K selection, but a {dynamic, iterative process} where the model updates its belief state after every glimpse, i.e., the selected regions are inserted into VLM's context in the form of visual tokens.

\begin{figure}[t]
    \centering
    \captionsetup[subfigure]{font=footnotesize, skip=2pt} 
    \begin{subfigure}[t]{0.49\linewidth}
        \centering
        \includegraphics[width=\linewidth]{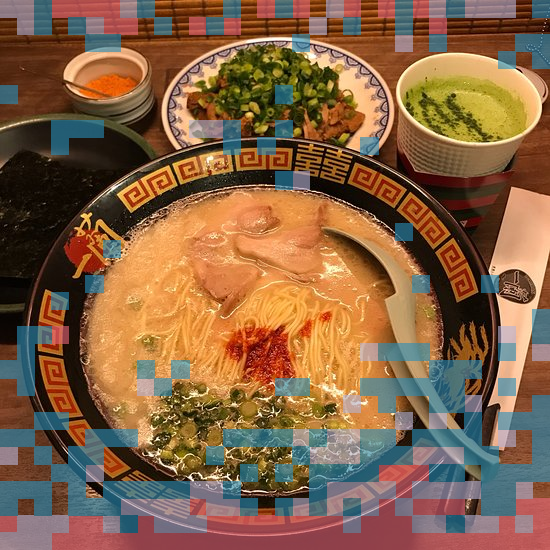}
        \subcaption{Regions selected by attention-based strategy.}
        \label{fig:22-a}
    \end{subfigure}\hfill
    \begin{subfigure}[t]{0.49\linewidth}
        \centering
        \includegraphics[width=\linewidth]{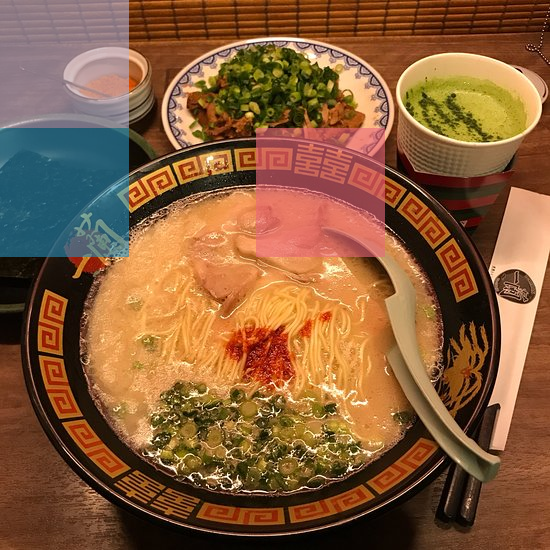}
        \subcaption{Regions selected by information-driven strategy.}
        \label{fig:22-b}
    \end{subfigure}

    \vspace{-0.4em} 
    \caption{A comparison of regions selected by different strategies.
    Details are provided in Appendix \ref{app: explanation for fig 2}.}
    \label{fig: 22nd example - motivation 2}
\end{figure}

These two theoretical insights directly shape the architecture of our AVP module (Section~\ref{sec: method 2}). Figure~\ref{fig: 22nd example - motivation 2} compares AVP-selected regions with those from attention-based method, which is further detailed in Appendix~\ref{app: explanation for fig 2}.

\subsection{{Identifying When to See: From Static Heuristics to Dynamic Attention Shifts}}
\label{sec: motivation for idea 3}

The fundamental goal of I-MCoT is to ground the VLM's reasoning in inserted fine-grained visual evidence. Therefore, the insertions should align with the model's fluctuating need for visual evidence across reasoning steps. However, existing methods rely on static triggers like newlines~\cite{icot}, failing to capture critical moments when the VLM actively seeks visual evidence to support subsequent generation.

To address this, we propose that the attention shift from textual to visual contexts (i.e., the text-to-vision attention shift) serves as a superior, dynamic indicator. Intuitively, a significant pivot in attention towards the visual modality suggests the VLM's cognitive demand for visual grounding. Unlike static triggers, monitoring these shifts allows for determining the precise ``when'' based on the model's real-time internal state.

We empirically validate this intuition through an in-depth analysis of the baseline model on the LLaVA-W benchmark (detailed in Appendix~\ref{app: motivation for idea 3}). Our investigation correlates the timing of visual insertions with model performance, revealing two key observations:
\textbf{(1) Correlation analysis:} Synchronizing visual insertions with salient text-to-vision attention shifts strongly correlates with better generation quality.
\textbf{(2) Group analysis:} This dynamic shift pattern distinguishes high-quality responses from low-quality ones.

In conclusion, although attention is unreliable as a decision rule for selection, its shift is still a \textbf{reliable diagnostic signal} revealing when the model is seeking visual information. Motivated by this, we present DAT, an attention-shift trigger that dynamically activates visual insertions (Section~\ref{sec: method 3}).

\begin{figure*}
    \vspace{-0em}
    \includegraphics[width=\linewidth]{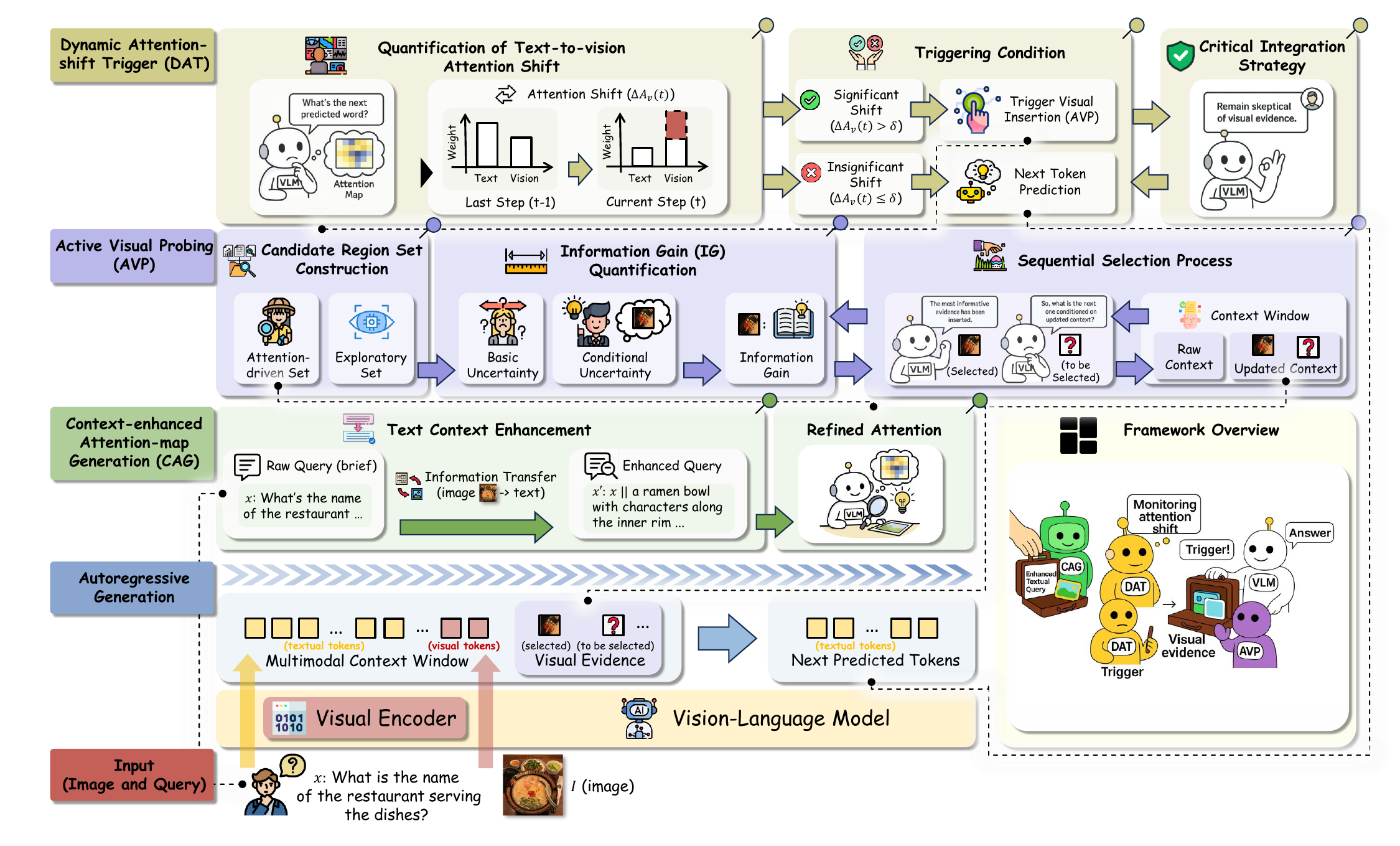}
    \vspace{-1cm}
    \caption{
    {An overview of the AIM-CoT framework. CAG initially enhances the query to refine attention distributions, thereby supporting reasoning and supplying reliable candidates for AVP. 
    During reasoning, AVP executes visual insertion by iteratively selecting the most informative regions as evidence conditioned on the current context. This insertion is triggered once DAT (the trigger) detects a significant text-to-vision shift in the VLM's internal attention state. See Appendix~\ref{app:fig_quick_guide} for a concise reading guide.
}
    }
    \label{fig: framework for avp}
    \vspace{-1em}
\end{figure*}

\section{AIM-CoT} \label{sec: method}

In this section, we begin by briefly reviewing the background of 
{vision-language reasoning}. Then, based on the motivations detailed in Section \ref{sec: motivations}, we present AIM-CoT. As a {training-free} framework, AIM-CoT models the interleaved-modal reasoning as an {information-foraging process} with three synergistic components: (1) \textbf{CAG}, which pre-processes the input to generate a fine-grained description, mitigating text-vision granularity imbalance. (2) \textbf{AVP}, which proactively selects regions that maximize information gain {whenever activated}. (3) \textbf{DAT}, which triggers AVP to {insert visual evidence} into the CoT precisely when the model's cognitive focus shifts from text to vision. {A visualization of AIM-CoT is shown in Figure \ref{fig: framework for avp}.}

It is important to distinguish AIM-CoT from attribution methods that rely on masking or deleting parts of the input image. AIM-CoT operates on a \textbf{frozen VLM} and adopts a  ``Trigger-Select-Insert'' paradigm: although sub-regions are integrated as visual evidence, the input context, including the original image is fully preserved. This ensures the VLM operates on a coherent, augmented context rather than a disrupted one \cite{igos++}.

\subsection{Preliminaries}
\label{sec:preliminary}


Due to space constraints, we defer the definitions of important concepts to Appendix \ref{app: concept intro}, including (1) the Vision-Language Models (VLMs), (2) the Context Window of a VLM, (3) Patches and Regions (i.e., Visual Tokenization), and (4) I-MCoT.

\subsection{Context-enhanced Attention-map Generation (CAG)}
\label{sec: method 1}

As revealed in Section \ref{sec: motivation for idea 1} and prior research~\cite{pyliu1}, attention is an unreliable basis for visual selection, particularly under severe text-vision granularity imbalance. 
Although AIM-CoT shifts from passive attention-based selection to proactive information-driven selection, we argue that refining the VLM's attention distribution remains meaningful for the interleaved-modal reasoning process. This is primarily for two reasons: 
(1) a more reliable attention distribution benefits  reasoning, since it serves as the basis of model generation;
and (2) crucially, the attention map serves as a source of candidate regions for subsequent selection (Section~\ref{sec: method 2}).

To this end, we propose \textbf{CAG} to improve the VLM's cross-attention distribution by mitigating the {text-vision granularity imbalance}. Specifically, before the VQA process begins, the VLM is elicited to carefully generate an explanatory description of the input image conditioned on the textual query. Rather than appending a caption, this process enriches the brief query with more semantic anchors that encode visual information within the image, thereby alleviating the granularity imbalance. Formally, this is expressed as follows:
\begin{align}
    \mathcal{D}_{\text{CAG}} &= \text{VLM}(I, x, \mathcal{P}_{\text{CAG}}), \\
    x' &= \text{concat}(x, \mathcal{D}_{\text{CAG}}),
\end{align}
where 
$\mathcal{P}_{\text{CAG}}$ is the prompt for generating the description. $x'$ is the updated textual query that integrates the description $\mathcal{D}_{\text{CAG}}$. 
A template of $\mathcal{P}_{\text{CAG}}$ is provided in Appendix~\ref{app: template of P_CAG}.

To ensure the output $\mathcal{D}_{\text{CAG}}$ serves as a reliable textual anchor while preventing error propagation to downstream modules, it is vital to mitigate potential hallucinations within it. To this end, the prompt $\mathcal{P}_{\text{CAG}}$ is meticulously designed with \textit{negative constraints}: it explicitly instructs the model to prioritize visible evidence and adopt a rigorous, cautious stance---skipping uncertain details rather than speculating. Experiments in Appendix~\ref{app:cag_reliability} validate the effectiveness of negative constraints.

\subsection{Active Visual Probing (AVP)}
\label{sec: method 2}

Drawing on the analyses in Section \ref{sec: motivation for idea 2}, we propose \textbf{AVP}, an information-driven mechanism designed to proactively select the regions with valuable information. Inspired by Information Foraging Theory (IFT), AVP operates through three systematic steps: (1) constructing a diverse candidate region set, (2) quantifying the information gain of each candidate region, and (3) executing a sequential, greedy selection process.

\noindent \textbf{Step 1: Candidate Region Set Construction.}
As introduced in Appendix \ref{app: concept intro}, each input image is partitioned into multiple regions with a fixed partitioning method. Although the partition for a given input image is deterministic, evaluating the information gain of each of these regions could be computationally prohibitive. Therefore, it is necessary to construct a candidate pool, which is a subset of the raw partition. 

We propose that although the candidate region set is a subset of the raw partition, it should remain sufficiently informative for subsequent selection. To this end, we derive candidate regions from diverse sources, covering both the VLM's current focus and potential blind spots, as follows:
\begin{itemize}
[topsep=0.5mm, partopsep=0pt, itemsep=0pt, leftmargin=10pt] 
    \item \textbf{Attention-driven Set ($C_{attn}$):} We select the top-$N$ regions from the CAG-enhanced cross-attention map ($A'$), as this subset is a real-time reflection of VLM's internal state in the latest reasoning step. Although the attention-based signals can inherently carry noise, analysis in Appendix \ref{app:avp_robustness} suggests that our subsequent information-driven selection effectively filters out non-informative high-attention regions (i.e., noise rejection).
    \item \textbf{Exploratory Set ($C_{exp}$):} To mitigate the ``tunnel vision'' of attention maps, we extract $M$ exploratory regions by uniformly sampling from all the partitioned regions. Despite its simplicity, this effective strategy outperforms complex alternatives (Appendix \ref{app: ablation study on C construction}) and provides essential regions overlooked by the attention-driven set $C_{attn}$ (Appendix \ref{app: proportion of sources}).
\end{itemize}
Formally, the total candidate set $C$ is constructed as follows:
\begin{align}
    C_{attn} &= \{R_1, R_2, \dots, R_N\},  \nonumber \\
    C_{exp}  &= \{R_{N+1}, R_{N+2}, \dots, R_{N+M}\},\nonumber  \\
    C        &= C_{attn} \cup C_{exp},\nonumber 
\end{align}
where each item denotes a specific image sub-region. For $i\leq N$, $R_i$ is the $i$-th most attended region in $A'$.

\noindent \textbf{Step 2: Information Gain Quantification.}
As analyzed in Section \ref{sec: motivation for idea 2}, IFT motivates us to identify the most informative regions. Therefore, it is necessary to quantify the information gain of each candidate region in $C$.

To quantify \textit{Information Gain (IG)} of a candidate region, we compare the VLM's information content before and after integrating this specific candidate (as visual tokens) into its current context. These pre- and post-insertion information contents are termed basic uncertainty and conditional uncertainty, respectively. Intuitively, they measure the VLM's uncertainty in predicting the next token (textual) during autoregression under these two states, respectively. 
Formally, \textit{Basic Uncertainty} ($U_B$) is defined as the entropy of the VLM's next-token distribution given the current context, and \textit{Conditional Uncertainty} ($U_{C,i}$) as the entropy after explicitly introducing candidate region $R_i$.
\begin{align}
   & U_B = H(Y | I, x, y_{<t}) \nonumber \\
        &= -\sum_{y \in V} P(y | I, x, y_{<t}) \log_2 P(y | I, x, y_{<t}), \\
   & U_{C,i} = H(Y | I, x, y_{<t}, R_i) \nonumber \\
            &= -\sum_{y \in V} P(y | I, x, y_{<t}, R_i) \log_2 P(y | I, x, y_{<t}, R_i), \nonumber 
\end{align}
where $V$ is the vocabulary, and $y_{<t}$ represents the generated tokens. The information gain is thus derived as:
\begin{equation}\label{eq:ig}
    IG(\{R_i\}) = U_B - U_{C,i}, \; i = 1, \dots, N+M.
\end{equation}
Although this calculation requires a forward pass, Appendix \ref{app: complexity of AVP} shows that AIM-CoT remains highly deployable, with an average inference time under $1.36\times$ that of the highly efficient baseline.

\noindent \textbf{Step 3: Sequential Selection Process.}
Finally, as analyzed in Section \ref{sec: motivation for idea 2}, drawing on the iterative nature of information foraging, where each step depends on previous knowledge, we frame region selection as a sequential trajectory. Specifically, in each step, we propose to select the region that provides the maximum information gain conditioned on the context updated in the last step, as outlined in Algorithm \ref{alg:greedy}. 
This is because this greedy algorithm (1) intuitively moves maximally toward providing the VLM with the most visual information at each step, and besides, (2) it has theoretical support: regarding such a subset selection problem, it yields an approximation to the global optimum (i.e., the maximum information gain) for the VLM. We provide comprehensive analyses of this theoretical property in Appendix \ref{app: proof for function F}.

\begin{algorithm}[t]
    \caption{Greedy Algorithm for Optimal Region Selection}
    \label{alg:greedy}
    \small
    \SetKwInOut{Input}{Input}
    \SetKwInOut{Output}{Output}
    
    \Input{Total candidate set $C$, Target selection size $K$}
    \Output{Optimal selection set $S$}
    
    $R^* \gets \varnothing$ \;
    \For{$k \gets 1$ \KwTo $K$}{
        Compute Basic Uncertainty: $U_B \gets H(Y | I, x, y_{<t}, R^*)$ \;
        \For{$R_i \in C \setminus R^*$}{
            Compute Conditional Uncertainty: $U_{C,i} \gets H(Y | I, x, y_{<t}, R^* \cup \{R_i\})$ \;
            Calculate Gain: $IG(\{R_i\}) \gets U_B - U_{C,i}$ \;
        }
        Select Best Region: $R_{next} \gets \operatorname{argmax}_{R_i \in C \setminus R^*} \{ IG(\{R_i\}) \}$ \;
        Update Selected Set: $R^* \gets R^* \cup \{R_{next}\}$ \;
    }
    $S \gets R^*$ \;
    \Return $S$
\end{algorithm}

\subsection{Dynamic Attention-shift Trigger (DAT)} \label{sec: method 3}

As identified in Section~\ref{sec: motivation for idea 3}, although attention remains an unreliable basis \textbf{for selection}, its shift serves as a reliable diagnostic signal \textbf{for triggering}. Motivated by this, we propose \textbf{DAT}, the trigger component of AIM-CoT. DAT tracks the VLM's text-to-vision attention shifts across the autoregressive steps, as they reflect the VLM's dynamic need for visual evidence. Then, based on the magnitude of each shift, DAT determines whether the visual evidence should be introduced via AVP at that step.

\paragraph{Quantification of Text-to-vision Attention Shift.}
The text-to-vision attention shift is determined by the difference in attention distributions between steps. Intuitively, it tracks the flow of attention from the text context in the previous step to the visual context in the current step. Furthermore, since the sum of attention scores over these two parts (i.e., text and vision) is fixed in the VLM's normalized attention map, it is sufficient to focus on the attention shift in the visual part. Formally, this can be quantified as follows:
\begin{equation}
    \Delta A_{vision}(t) = A_{vision}(t) - A_{vision}(t-1),
    \label{eq:attention_shift}
\end{equation}
where $A_{vision}(t)$ and $A_{vision}(t-1)$ represent the sum of attention scores allocated to all the visual tokens in the context by the currently predicted token (the $t$-th token) and the preceding token (the $(t-1)$-th token), respectively.

\paragraph{Triggering Condition.}
DAT determines if a given shift $\Delta A_{vision}(t)$ is significant enough to warrant the provision of visual evidence via a qualitative criterion. In the criterion, a hyper-parameter $\delta \in \mathbb{R}$ is introduced to serve as the threshold. The AVP module is triggered to perform visual insertion after the $t$-th generation step if and only if $\Delta A_{vision}(t) > \delta$. 

In Appendix~\ref{app: sensitivity analysis of delta}, we provide a detailed sensitivity analysis of $\delta$ and introduce an adaptive thresholding strategy to set $\delta$ automatically.

\paragraph{Critical Integration Strategy.}
A potential challenge inherent to the I-MCoT paradigm is the difficulty in eliminating noise that visual evidence might introduce. To address this, DAT does not naively trigger the visual evidence insertion. Instead, it employs a critical integration mechanism: the inserted regions are accompanied by a \textbf{safety instruction}, which cues the VLM to treat the visual evidence as ``supplementary references'', encouraging the model to verify semantic consistency with the existing textual context. This effectively filters out irrelevant or hallucinated visual cues.

We validate the effectiveness of this safety instruction in Appendix~\ref{app:dat_safety}.

\begin{table*}[htbp]
    \centering
    \caption{Performance comparison results on three VQA benchmarks and four backbones. The best performances are shown in bold. We report Accuracy (ACC.) for M3CoT and ScienceQA, and ROUGE-L for LLaVA-W.}
    \label{tab: performance_chameleon}
    
    \footnotesize
    \setlength{\tabcolsep}{0pt}
    
    \begin{minipage}[t]{0.49\textwidth}
        \centering
        \begin{tabular*}{\linewidth}{@{\extracolsep{\fill}}lcccccc} 
            \toprule
            \multirow{2}{*}{Method} & \multicolumn{2}{c}{M3CoT} & \multicolumn{2}{c}{ScienceQA} & \multicolumn{2}{c}{LLaVA-W} \\ 
            \cmidrule(lr){2-3} \cmidrule(lr){4-5} \cmidrule(lr){6-7}
            & 0-shot & 1-shot & 0-shot & 1-shot & 0-shot & 1-shot \\  
            \midrule
            
            \multicolumn{7}{c}{\cellcolor{gray!10}\textbf{Chameleon-7B}} \\ 
            \midrule
            No-CoT & 29.1 & 28.4 & 47.7 & 48.5 & 13.1 & 23.9 \\
            DDCoT & 28.6 & 29.8 & 49.8 & 49.2 & 20.2 & 23.1 \\
            MMCoT & 28.5 & 30.6 & 49.0 & 50.7 & 20.4 & 20.6 \\
            CCoT & 29.4 & 31.4 & 50.2 & 51.3 & 22.1 & 24.5 \\
            SCAFFOLD & 29.6 & 31.1 & 48.5 & 47.5 & 21.7 & 24.7 \\
            ICoT & 29.8 & 32.3 & 51.0 & 53.4 & 25.2 & 27.6 \\ 
            \textbf{AIM-CoT (ours)} &
            \textbf{31.4}$^{\ddagger}$ & \textbf{32.8}$^{\dagger}$ &
            \textbf{53.1}$^{\ddagger}$ & \textbf{54.5}$^{\dagger}$ &
            \textbf{29.8}$^{\ddagger}$ & \textbf{32.0}$^{\ddagger}$ \\ 
            \textcolor{green}{\textit{Improv.}} & \textcolor{green}{\textit{+5.4\%}} & \textcolor{green}{\textit{+1.5\%}} & \textcolor{green}{\textit{+4.1\%}} & \textcolor{green}{\textit{+2.1\%}} & \textcolor{green}{\textit{+18.3\%}} & \textcolor{green}{\textit{+15.9\%}} \\
            
            \midrule 
            \multicolumn{7}{c}{\cellcolor{gray!10}\textbf{Janus-Pro-7B}} \\ 
            \midrule
            No-CoT & 36.5 & 37.8 & 53.9 & 61.2 & 29.8 & 30.9 \\ 
            DDCoT & 36.8 & 38.6 & 53.2 & 61.6 & 29.0 & 30.4 \\ 
            MMCoT & 35.2 & 36.6 & 50.8 & 58.0 & 28.2 & 28.9 \\ 
            CCoT & 36.4 & 38.2 & 54.1 & 61.5 & 27.8 & 31.5 \\ 
            SCAFFOLD & 36.1 & 38.0 & 52.4 & 60.5 & 29.7 & 30.8 \\ 
            ICoT & 37.6 & 39.4 & 55.1 & 62.5 & 32.5 & 33.6 \\ 
            \textbf{AIM-CoT (ours)} &
            \textbf{39.7}$^{\ddagger}$ & \textbf{41.5}$^{\ddagger}$ &
            \textbf{56.9}$^{\ddagger}$ & \textbf{64.9}$^{\ddagger}$ &
            \textbf{35.5}$^{\ddagger}$ & \textbf{36.6}$^{\ddagger}$ \\ 
            \textcolor{green}{\textit{Improv.}} & \textcolor{green}{\textit{+5.6\%}} & \textcolor{green}{\textit{+5.3\%}} & \textcolor{green}{\textit{+3.3\%}} & \textcolor{green}{\textit{+3.8\%}} & \textcolor{green}{\textit{+9.2\%}} & \textcolor{green}{\textit{+8.9\%}} \\ 
            \bottomrule
        \end{tabular*}
    \end{minipage}
    \hfill
    \begin{minipage}[t]{0.49\textwidth}
        \centering
        \begin{tabular*}{\linewidth}{@{\extracolsep{\fill}}lcccccc} 
            \toprule
            \multirow{2}{*}{Method} & \multicolumn{2}{c}{M3CoT} & \multicolumn{2}{c}{ScienceQA} & \multicolumn{2}{c}{LLaVA-W} \\ 
            \cmidrule(lr){2-3} \cmidrule(lr){4-5} \cmidrule(lr){6-7}
            & 0-shot & 1-shot & 0-shot & 1-shot & 0-shot & 1-shot \\  
            \midrule
            
            \multicolumn{7}{c}{\cellcolor{gray!10}\textbf{Qwen2-VL-7B}} \\ 
            \midrule
            No-CoT & 43.6 & 45.4 & 56.3 & 64.4 & 32.7 & 33.5 \\
            DDCoT & 42.6 & 45.7 & 55.2 & 64.9 & 31.2 & 32.8 \\
            MMCoT & 40.1 & 42.5 & 51.3 & 58.3 & 30.7 & 31.4 \\
            CCoT & 43.3 & 44.1 & 56.4 & 63.8 & 29.4 & 33.9 \\
            SCAFFOLD & 41.7 & 44.9 & 53.7 & 62.5 & 31.8 & 33.1 \\
            ICoT & 44.1 & 46.0 & 56.8 & 65.4 & 34.2 & 35.7 \\
            \textbf{AIM-CoT (ours)} &
            \textbf{44.7}$^{\dagger}$ & \textbf{46.6}$^{\dagger}$ &
            \textbf{57.4}$^{\dagger}$ & \textbf{66.3}$^{\dagger}$ &
            \textbf{36.3}$^{\dagger}$ & \textbf{37.3}$^{\dagger}$ \\ 
            \textcolor{green}{\textit{Improv.}}  & \textcolor{green}{\textit{+1.4\%}} & \textcolor{green}{\textit{+1.3\%}} & \textcolor{green}{\textit{+1.1\%}} & \textcolor{green}{\textit{+1.4\%}} & \textcolor{green}{\textit{+6.1\%}} & \textcolor{green}{\textit{+4.5\%}} \\ 
            
            \midrule
            \multicolumn{7}{c}{\cellcolor{gray!10}\textbf{Qwen2.5-VL-32B}} \\ 
            \midrule
            No-CoT & 55.2 & 56.8 & 73.5 & 77.2 & 40.5 & 41.8 \\
            DDCoT & 54.8 & 57.5 & 72.9 & 77.8 & 39.6 & 41.3 \\
            MMCoT & 53.4 & 54.7 & 69.8 & 74.5 & 38.9 & 39.2 \\
            CCoT & 55.3 & 56.9 & 73.4 & 77.1 & 37.8 & 42.4 \\
            SCAFFOLD & 54.2 & 56.7 & 71.6 & 76.3 & 40.1 & 41.5 \\
            ICoT & 56.9 & 59.1 & 75.1 & 79.2 & 43.4 & 44.7 \\
            \textbf{AIM-CoT (ours)} &
            \textbf{58.7}$^{\ddagger}$ & \textbf{61.2}$^{\ddagger}$ &
            \textbf{76.8}$^{\ddagger}$ & \textbf{81.3}$^{\dagger}$ &
            \textbf{46.5}$^{\ddagger}$ & \textbf{49.1}$^{\ddagger}$ \\ 
            \textcolor{green}{\textit{Improv.}}  & \textcolor{green}{\textit{+3.2\%}} & \textcolor{green}{\textit{+3.6\%}} & \textcolor{green}{\textit{+2.3\%}} & \textcolor{green}{\textit{+2.7\%}} & \textcolor{green}{\textit{+7.1\%}} & \textcolor{green}{\textit{+9.8\%}} \\ 
            \bottomrule
        \end{tabular*}
    \end{minipage}

    \vspace{2pt}
    \scriptsize
    \raggedright
    \textit{Statistical significance:}
    $^{\dagger}p<0.05$, $^{\ddagger}p<0.01$ for AIM-CoT compared with the second-best method under the same backbone/setting
    (McNemar for ACC. on M3CoT/ScienceQA; Wilcoxon signed-rank for ROUGE-L on LLaVA-W). Full results are reported in Appendix \ref{app:statistical_test}.
\end{table*}

\section{Experiments}

\subsection{Evaluation Setup} \label{sec: benchmarks}

\textbf{Benchmark.} We evaluate AIM-CoT on three widely used VQA benchmarks: M3CoT~\cite{m3cot}, ScienceQA~\cite{sqa}, and LLaVA-W~\cite{llava-w}. Detailed descriptions are provided in Appendix~\ref{app: introduction to benchmarks}.

\textbf{Baselines.} We compare against several text-only baselines, including vanilla VLM without CoT (No-CoT), DDCoT~\cite{ddcot}, MMCoT~\cite{mmcot}, CCoT~\cite{ccdot}, and SCAFFOLD~\cite{scaffold}. On top of these, the leading I-MCoT framework ICoT~\cite{icot}, is also included. More details are available in Appendix~\ref{app: introduction to baseline models}. When possible, we directly report figures from prior work.

\textbf{Backbones.} The implementation is built on four mainstream VLM backbones with different architectures and scales: early-fusion Chameleon-7B~\cite{chameleon} and Janus-Pro-7B~\cite{janus}, and late-fusion Qwen2-VL-7B~\cite{qwen} and Qwen2.5-VL-32B~\cite{qwen2.5}. We conduct experiments in both 0- and 1-shot settings, using the prompt template from the open-source ICoT implementation~\cite{icot}.

\textbf{Hyper-parameter.} Hyper-parameter settings are listed in Appendix~\ref{app: hyper-parameter settings}.

\begin{table}[htbp]
    \centering
    \caption{Ablation study of AIM-CoT conducted on Chameleon-7B under 0-shot setting.}
    \label{tab: ablation study (ours)}
    \resizebox{\columnwidth}{!}{%
        \begin{tabular}{@{}lcccc@{}}
            \toprule
            Dataset & AIM-CoT & w/o CAG & w/o AVP & w/o DAT \\
            \midrule
            M3CoT (ACC.) & 31.4 & 30.5 (\textcolor{red}{-0.9}) & 30.6 (\textcolor{red}{-0.8}) & 30.8 (\textcolor{red}{-0.6}) \\
            ScienceQA (ACC.) & 53.1 & 52.8 (\textcolor{red}{-0.3}) & 52.3 (\textcolor{red}{-0.8}) & 52.7 (\textcolor{red}{-0.4}) \\
            LLaVA-W (ROUGE-L) & 29.8 & 26.8 (\textcolor{red}{-3.0}) & 26.2 (\textcolor{red}{-3.6}) & 27.3 (\textcolor{red}{-2.5}) \\
            \bottomrule
        \end{tabular}%
    }
\end{table}

\subsection{Performance Comparison}

Table \ref{tab: performance_chameleon} reports the experimental results. AIM-CoT consistently outperforms all baselines (both text-only and I-MCoT) across all benchmarks and backbones, under both 0- and 1-shot settings. This demonstrates the strong advantage of AIM-CoT for complex vision-language reasoning (besides, the analyses in Appendix \ref{app: complexity of AVP} show that AIM-CoT maintains good deployability). 

More specifically, both I-MCoT methods (AIM-CoT and ICoT) surpass the text-only baselines, confirming the efficacy of explicit incorporation of visual evidence. Moreover, although AIM-CoT and ICoT follow the same I-MCoT paradigm, AIM-CoT achieves stronger performance. This gain comes from three key improvements: (1) AIM-CoT uses CAG to directly refine the VLM's cross-attention distribution, instead of grounding in the unreliable signal used by ICoT; (2) it replaces fragile attention-based heuristics with information-driven selection (AVP); and (3) it replaces a static trigger with a dynamic and more precise trigger (DAT).

Finally, the performance gains vary across backbones. In Appendix \ref{sec:performance_variance_analysis}, we analyze this effect in detail and attribute it to the interaction between model architecture and model scale.

\subsection{Ablation Study}\label{sec: ablation study (ours)}

In this section, we conduct the ablation study to verify the efficacy of each component within AIM-CoT. The detailed settings are as follows:
\begin{itemize}[topsep=0.5mm, partopsep=0pt, itemsep=0pt, leftmargin=10pt]
    \item \textbf{w/o CAG}: The VLM operates under the image $I$ and the raw query $x$, instead of the CAG-enhanced query $x'$.
    \item \textbf{w/o AVP}: The information-driven selection (AVP) is replaced by attention-based selection.
    \item \textbf{w/o DAT}: DAT is substituted with a static trigger (i.e., visual evidence is inserted whenever the VLM outputs a newline character).
\end{itemize}
The results on Chameleon-7B are shown in Table \ref{tab: ablation study (ours)} (more results and analyses on other backbones are deferred to Appendix~\ref{app:extensive ablation studies}). 
First, refining the VLM's attention distribution via CAG can strongly improve VLM's performance on VQA tasks. This is because the attention is not only the basis of model generation but also one source of candidate evidence.
Furthermore, (1) even without CAG, AIM-CoT still significantly outperforms ICoT (Table~\ref{tab: performance_chameleon}); and (2) removing AVP and DAT causes a larger performance drop than removing CAG. These results highlight the importance of tackling the \textit{what to see} and \textit{when to see it} questions. In response to these two issues, we propose AVP and DAT, respectively.

To provide a comprehensive understanding of AIM-CoT, we go beyond merely validating component efficacy. In Sections~\ref{app:semantic_relevance},~\ref{app:human_alignment_gpt4v}, and Appendix \ref{app:qualitative_case_IG_vs_AS}, we present both quantitative and visualized qualitative analyses of AVP, elucidating why information-driven regions are more effective than attention-based regions.

\subsection{Interplay between CAG and AVP} 
\label{sec: analysis of interplay between CAG and AVP}

In this section, we investigate the interaction between CAG and AVP. This is achieved by sequentially adding them to a basic model (BM) (i.e., AIM-CoT stripped of all its components). 

As shown in Table \ref{tab: ablation study (baseline)}, while CAG and AVP yield individual gains, their combination exhibits a clear phenomenon of \textbf{super-additivity}, i.e., the joint improvement consistently exceeds the sum of their separate contributions. Specifically, on M3CoT, the joint gain ($+1.0\%$) significantly surpasses the linear sum of individual gains ($0.3\% + 0.4\% = 0.7\%$). This trend holds across all benchmarks (e.g., $+1.7\% > 1.4\%$ on ScienceQA). 

This non-linear boost validates the interplay between the modules: CAG refines the VLM's attention distribution, which amplifies AVP's effectiveness; the AVP, in turn, precisely mines the salient visual evidence from the attention-driven candidates ($C_{attn}$) enhanced by CAG.

\begin{table}[t]
    \centering
    \caption{Ablation study of the baseline model (BM) on Chameleon-7B under 0-shot setting.}
    \label{tab: ablation study (baseline)}
    \resizebox{\columnwidth}{!}{%
        \begin{tabular}{@{}lcccc@{}}
            \toprule
            Dataset & BM & BM w/ CAG & BM w/ AVP & BM w/ CAG, AVP \\
            \midrule
            M3CoT (ACC.) & 29.8 & 30.1 (\textcolor{green}{+0.3}) & 30.2 (\textcolor{green}{+0.4}) & 30.8 (\textcolor{green}{+1.0}) \\
            ScienceQA (ACC.) & 51.0 & 51.5 (\textcolor{green}{+0.5}) & 51.9 (\textcolor{green}{+0.9}) & 52.7 (\textcolor{green}{+1.7}) \\
            LLaVA-W (ROUGE-L) & 25.2 & 25.8 (\textcolor{green}{+0.6}) & 26.4 (\textcolor{green}{+1.2}) & 27.3 (\textcolor{green}{+2.1}) \\
            \bottomrule
        \end{tabular}%
    }
\end{table}

\subsection{Quantitative Analysis: Semantic Relevance}
\label{app:semantic_relevance}

We further examine whether the visual regions selected by AIM-CoT match human-centered perception and the needs of the task. Our premise is that informative evidence should correspond to distinct and complete semantic entities rather than repetitive background patterns. We therefore use the Segment Anything Model (SAM) \cite{sam} as a proxy to assess whether the selected regions align with meaningful object concepts.

We conduct this evaluation on M3CoT and ScienceQA, randomly sampling 500 instances from each benchmark. Results are averaged over the four backbones: Chameleon-7B, Janus-Pro-7B, Qwen2-VL-7B, and Qwen2.5-VL-32B. To build a reference for visually relevant evidence, we use SAM to generate segmentation masks for the key entities mentioned in the question and ground-truth answer. We then define the \textbf{Semantic Hit Rate (SHR)} as the percentage of instances in which the selected regions substantially overlap with the corresponding SAM masks, measured by IoU greater than 0.5.

\begin{table}[t]
\centering
\caption{Comparison of SHR on M3CoT and ScienceQA (subset of 500 samples each). The results are averaged across four backbones.}
\label{tab:semantic_hit_rate}
\resizebox{\columnwidth}{!}{%
\begin{tabular}{lccc}
\toprule
\multirow{2}{*}{Method} & \multirow{2}{*}{Selection Strategy} & \multicolumn{2}{c}{SHR $\uparrow$} \\
\cmidrule(lr){3-4}
 & & M3CoT & ScienceQA \\
\midrule
ICoT \cite{icot} & Attention-driven Top-K & 18.7\% & 24.3\% \\
\textbf{AIM-CoT (Ours)} & \textbf{Info. Gain-driven AVP} & \textbf{65.2\%} & \textbf{71.8\%} \\
\bottomrule
\end{tabular}%
}
\end{table}

Table \ref{tab:semantic_hit_rate} shows that the attention-driven Top-K strategy used by ICoT struggles to recover semantically relevant evidence, with SHR dropping to 18.7\% on M3CoT. This supports the observation that raw attention maps often drift toward high-contrast background patterns or generic salient objects, rather than the specific evidence required for reasoning. In contrast, AIM-CoT achieves substantially higher SHR scores, reaching 65.2\% on M3CoT and 71.8\% on ScienceQA. These results indicate that our active information-seeking paradigm filters out distractors and selects regions that are both structurally coherent and semantically aligned with the question, making them closer to human judgments of relevant evidence.

\subsection{Quantitative Analysis: Alignment with Human Intuition}
\label{app:human_alignment_gpt4v}

We next evaluate whether the visual regions selected by AIM-CoT better align with human intuition than those selected by the attention-driven baseline. Following prior work, we use GPT-4v \cite{gpt4v} as an expert proxy judge, as strong multimodal judges have been shown to correlate well with human preferences \cite{judge_1,judge_2,judge_3,judge_4,judge_5}.

We evaluate 500 randomly sampled instances from the M3CoT and LLaVA-W benchmarks. For each instance, GPT-4v is given the textual query, the original image, and two anonymized sets of selected regions: one produced by ICoT (Top-K) and the other by AIM-CoT (AVP). We randomize the order of the two sets to eliminate position bias. GPT-4v then compares them along three criteria: semantic relevance, object completeness, and overall helpfulness. The evaluation prompt is shown in Table \ref{tab:gpt4v_prompt_box}.

\begin{table}[t]
\centering
\caption{The core prompt used for GPT-4v blind pairwise comparison.}
\label{tab:gpt4v_prompt_box}
\fbox{
\begin{minipage}{0.92\linewidth}
\small
\textbf{System Prompt:} You are an expert judge. Compare two sets of image crops (Set A and Set B) extracted from the Original Image. Decide which set is more helpful for a human to answer the Question. \\
\textbf{Criteria:} \\
1. \textbf{Relevance:} Does the set contain the specific objects or details mentioned in the question?\\
2. \textbf{Completeness:} Are the objects complete, or are they meaningless background noise/fragments?\\
3. \textbf{Helpfulness:} Which set would better help a human answer the question without seeing the full image?\\
\textbf{Output:} 'Set A is better', 'Set B is better', or 'Tie'.
\end{minipage}
}
\end{table}

\begin{table}[t]
\centering
\caption{GPT-4v preference rates on 500 samples from M3CoT and LLaVA-W. ``Win'' denotes that AIM-CoT provides better visual evidence than ICoT. ``Tie'' indicates equal quality.}
\label{tab:gpt4v_win_rate}
\resizebox{\columnwidth}{!}{
\begin{tabular}{lccc}
\toprule
Benchmark & \textbf{AIM-CoT Win} & {Tie} & {ICoT Win} \\
\midrule
M3CoT (Reasoning) & \textbf{76.4\%} & 16.2\% & 7.4\% \\
LLaVA-W (In-the-Wild) & \textbf{81.2\%} & 13.6\% & 5.2\% \\
\bottomrule
\end{tabular}
}
\end{table}

As shown in Table \ref{tab:gpt4v_win_rate}, AIM-CoT is strongly preferred across both benchmarks. It wins on 76.4\% of M3CoT samples and 81.2\% of LLaVA-W samples, while only a small fraction favor ICoT. These results suggest that AIM-CoT more consistently identifies complete and semantically informative evidence that matches human intuition. GPT-4v's qualitative judgments further indicate that ICoT often focuses on noisy regions, whereas AIM-CoT more reliably localizes the full objects needed for reasoning.

\section{Conclusion}
In this paper, we propose AIM-CoT, a novel I-MCoT framework that aims to frame the construction of interleaved-modal CoT as an active information-foraging process. Existing methods' static triggers fail to capture the dynamic needs of the VLM for visual information in complex reasoning, and their attention-based selectors depend heavily on the VLM's attention, which can be unreliable (for selection).
In response to these challenges, AIM-CoT dynamically monitors the VLM's cognitive need for fine-grained visual evidence for subsequent generation, and accordingly, selects salient visual evidence in an information-driven manner.
Extensive experiments demonstrate that AIM-CoT outperforms the leading methods across three benchmarks and four VLM backbones.


\section{Acknowledgements}

This work was partially supported by the Guangdong Basic and Applied Basic Research Foundation under Grant No. 2024A1515011949 and 2026A1515011672, the Shenzhen Science and Technology Program under Grant No. GXWD20231130110308001 and JCYJ20250604145617023.



\bibliography{custom}

\appendix

\section{Experimental Setup and Reproducibility}

\subsection{Benchmarks} \label{app: introduction to benchmarks}

\textbf{M3CoT} \cite{m3cot} is a novel multimodal CoT benchmark, which introduces complex, multi-step problems across science, mathematics, and commonsense domains, comprising 11,459 samples in total. M3CoT is characterized by succinct textual queries (<15 tokens on average) paired with intricate problems. This inherent text-vision imbalance makes it an ideal platform to validate the efficacy of our proposed CAG in mitigating this issue and the superiority of AVP in proactively selecting the salient visual regions.

\textbf{ScienceQA} \cite{sqa} is a popular benchmark for multiple-choice question answering with explanations on scholarly articles, comprising over 100,000 context-question-answer triples to address data scarcity in scientific machine reading comprehension. 

\textbf{LLaVA-Bench In-the-Wild} (LLaVA-W) \cite{llava-w} is a challenging open-ended benchmark designed to evaluate the real-world capabilities of VLMs by mimicking the unpredictability of real-world scenarios. The answers generated by GPT-4v \cite{gpt4v} serve as the labels. LLaVA-W is exceptionally well-suited for evaluating the capability of our proposed framework to address complex, open-ended problems by generating a multimodal CoT, attending to salient regions within the image, and meticulously parsing the query.

\subsection{Baselines} \label{app: introduction to baseline models}

\textbf{No-CoT} prompts the VLM to answer questions directly based on the input query and image. In the 1-shot setting, an example containing the query, image, and corresponding answer is attached, which follows the paradigm of in-context learning~\cite{wqwang1,wqwang2,wqwang3}.

\textbf{DDCoT} \cite{ddcot} deconstructs a multimodal problem into reasoning and recognition sub-questions, uses negative-space prompting to identify and fill visual information gaps with external models, and then integrates all information for a final joint reasoning step to generate rationales. 

\textbf{MMCoT} \cite{mmcot} first generates a rationale from fused language and vision inputs, and then uses this rationale along with the original multimodal data to infer the final answer.

\textbf{CCoT} \cite{ccdot} first prompts the VLM to generate a scene graph from an image and then uses it as an intermediate reasoning step to produce the final response.

\textbf{SCAFFOLD} \cite{scaffold} promotes vision-language coordination in the VLM by overlaying a dot matrix with coordinates onto an image, which then serves as a visual anchor that can be explicitly referenced in the textual prompt. 

\textbf{ICoT} \cite{icot} leverages the attention maps of the VLM to select relevant patches from the input image and insert them into the reasoning process, thereby generating sequential steps of paired visual and textual rationales.

\subsection{Template of $\mathcal{P}_{CAG}$} \label{app: template of P_CAG}

\begin{figure*}[t]
\centering
\begin{minipage}{\textwidth}
 \includegraphics[width=\textwidth]{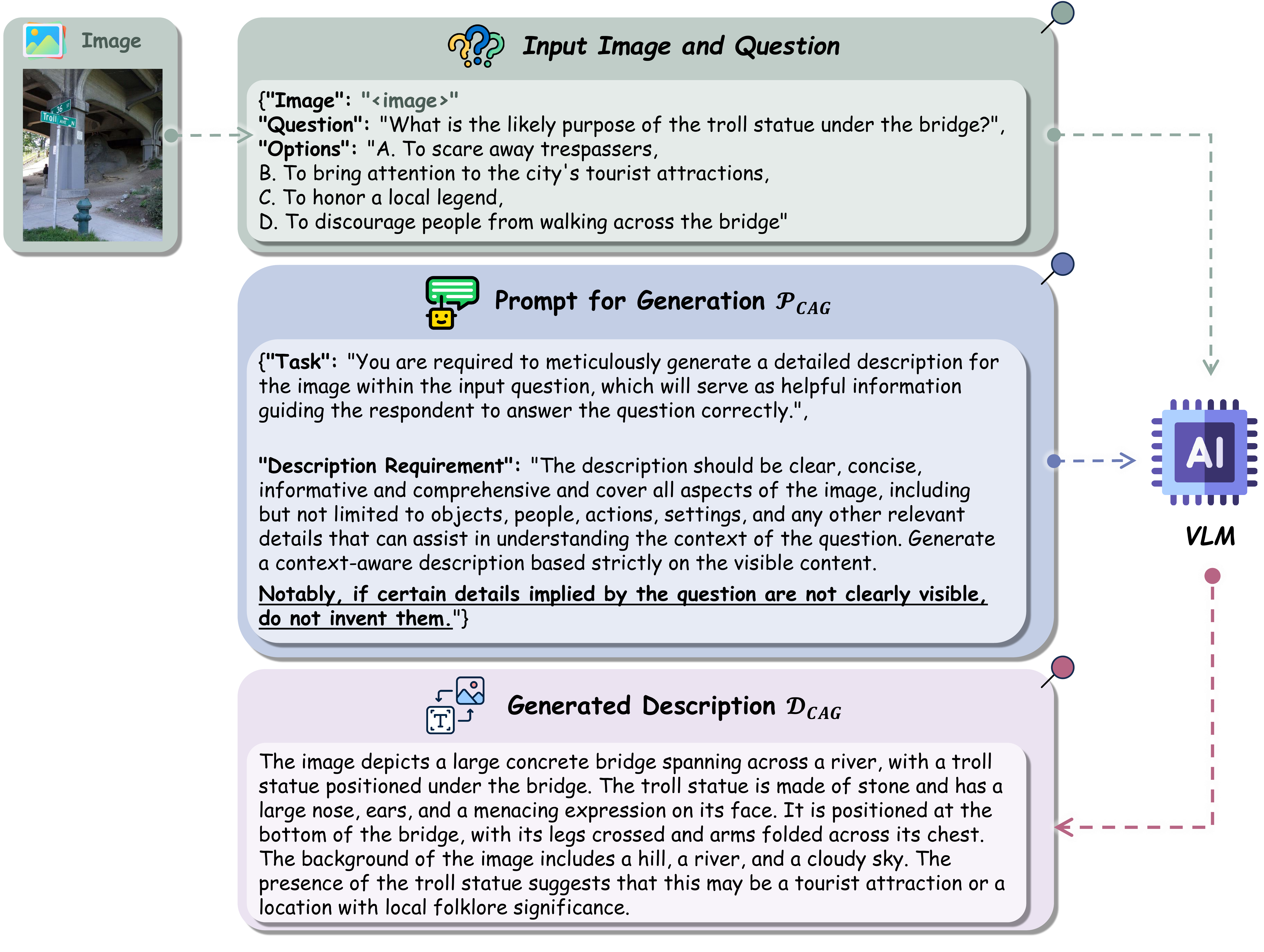}
\caption{An illustration of the entire process of context enhancement by the CAG module, using problem physical-commonsense-1398 from the M3CoT benchmark as an example. This showcases both the template and usage of $\mathcal{P}_{CAG}$.}
\label{fig: template of P_CAG}
\end{minipage}
\end{figure*}

Figure \ref{fig: template of P_CAG} provides an intuitive example showing the template of $\mathcal{P}_{CAG}$ and how it is used to prompt the VLM to carefully generate a guiding description for the input image. In particular, for multiple-choice questions, such as those in M3CoT and ScienceQA, we prepend the following brief explanation to $\mathcal{P}_{CAG}$ to aid the VLM in better understanding its designated task: ``\textit{This is a multiple-choice question. The question is based on the image provided.}'' 

Then, the cross-attention weight matrix based on the enhanced context $x'$ and $I$ can be obtained as follows:
\begin{align}
    A'=\text{softmax}(\frac{(H_TW^Q)(H_VW^K)^T}{\sqrt{d_K}}),
\end{align}
where $H_T\in \mathds{R}^{n_T\times d}, H_V\in \mathds{R}^{n_V\times d}$ are the hidden states of the textual and visual input, respectively. $W^Q, W^K$ are the weight matrices of the linear transformation layers for query and key, respectively.

\subsection{Setting of Hyper-parameters} \label{app: hyper-parameter settings}

The hyper-parameter settings are listed in Table \ref{tab: hyper-parameter}.

\begin{table}[htbp]
\centering
\caption{Hyper-parameter settings across three datasets.}
\label{tab: hyper-parameter}
\resizebox{\columnwidth}{!}{%
    \begin{tabular}{@{}cccc@{}}
    \toprule
    Parameter & M3CoT & ScienceQA & LLaVA-W \\ \midrule
    $N_C$ & 8 & 8 & 6 \\
    $K$ & 3 & 3 & 3 \\
    $N$ & 4 & 4 & 2 \\
    $M$ & 4 & 4 & 1 \\
    Region size for AVP $s_r$ (grid) & 1 & 1 & 1 \\
    Grid size for AVP $s_g$ & 4 & 4 & 4 \\
    $\delta$ & 0.5 & 0.2 & 0.2 \\
    \midrule
    Temperature & 0.7 & 0.7 & 0.7\\
    Do sample & True & True & True \\
    Top\_p & 0.9 & 0.9 & 0.9 \\
    Repetition\_penalty & 1.2 & 1.2 & 1.2\\
    Min\_new\_tokens & 32 & 32 & 32 \\
    Max\_new\_tokens & 512 & 1024 & 1024\\
    \bottomrule
    \end{tabular}%
}
\end{table}

\section{Supplementary Important Definitions}
\label{app: concept intro}

\paragraph{Vision-Language Model.}
A VLM typically fuses a vision encoder for preprocessing visual input and a generative language model, which jointly enable it to respond in a human-like manner. On the efficiency side, recent work on diffusion MLLMs explores phase mapping and sparsity-aware truncation to reduce generative inference cost~\cite{zhao2026resilphase,zhao2026seeingendstepzero}:
\begin{align}\label{eq: w/o CoT}
    answer = \text{VLM}(I,x),
\end{align}
where $I$ is the image and $x$ the query.

\paragraph{Context Window of a VLM.}
An autoregressive generation of a VLM is conditioned on (1) the user's initial input and (2) its interaction with the user. A context window is a structured representation of this conditional information.

Specifically, a context window starts with the user's multimodal input question, including a textual query $x$ and a paired image $I$.
Built upon this, the VLM continuously extends the initial context window by iteratively incorporating each reasoning step and its corresponding visual evidence (if any) into this context, once they are acquired.
This ensures that the prior part of the context window can consistently yet iteratively influence the subsequent generation and the final response.

In implementation, this multimodal information, whether textual or visual context, is incorporated into the context window in the form of tokens.

\paragraph{Visual Tokenization: Patches and Regions.}
In VLM's visual encoder preprocessing, the input image $I$ is transformed into a sequence of patches and visual tokens. First, $I$ is divided into a fixed number of non-overlapping and equal-sized \textbf{patches}. Each patch is a local fragment of $I$, which is then projected into a corresponding visual token, serving as the fundamental atomic unit for the model's attention mechanism.

A \textbf{region} is a spatial crop comprised of multiple spatially contiguous (i.e., neighboring) patches. Compared to a single patch, a region covers a larger area to capture higher-level, concrete semantic information that a single atomic unit (i.e., patch) lacks. Compared to the entire image $I$, a selected region enables the provision of finer-grained visual details.

\paragraph{Mapping from Patches to Regions.}
For a given input image, and given hyperparameters $s_r$ and $s_g$, the mapping from patches to regions is deterministic. However, this relies on an intermediate concept, the ``grid''. Specifically,
in the AVP module, the input image is first divided into $s_g\times s_g$ grids according to the set ``Grid size'' $s_g$. Each grid is comprised of multiple patches. For example, Chameleon \cite{chameleon} divides the input image into $32 \times 32 = 1024$ patches. Therefore, each grid contains $(32/s_g) \times (32/s_g) = 1024/{s_g}^2$ patches.
Furthermore, each region is a group of spatially continuous grids, consisting of $s_r\times s_r$ grids.

\paragraph{Interleaved-Modal CoT (I-MCoT).} 
Built upon a VLM backbone, the I-MCoT paradigm fundamentally relies on two key components: {selection} and {triggering}. Compared to direct response (Equation~\ref{eq: w/o CoT}), the trigger mechanism temporarily pauses the textual autoregressive process when specific conditions are met. During this suspension, the I-MCoT method selects salient visual evidence from the input image, which is then tokenized and concatenated into VLM's context to inform and support subsequent reasoning.

\section{Supplementary Details Regarding Motivation in Section~\ref{sec: motivation for idea 1}}

\subsection{Experiment for Sufficiency Check}
\label{app:supplementary for sufficiency check in motivation 1}

In the sufficiency check in Section~\ref{sec: motivation for idea 1}, we assess the impact of the most attended regions on the baseline model (i.e., ICoT~\cite{icot}) by masking them out.

\paragraph{Experimental Setup.}
We conduct experiments with four VLM backbones (Chameleon-7B~\cite{chameleon}, Janus-Pro-7B~\cite{janus}, Qwen2-VL-7B~\cite{qwen}, and Qwen2.5-VL-32B~\cite{qwen2.5}) and report results averaged across all backbones. We adopt the zero-shot setting.

\paragraph{Experimental Results.}
The results are shown in Table~\ref{tab: performance degradation when masked}. As $K_{\text{mask}}$ increases, performance drops slightly, but the decline remains small. Even when $K_{\text{mask}}=50$, the model does not exhibit a substantial degradation in performance.

\begin{table}[htbp]
    \centering
    \caption{Performance difference of the baseline model (ICoT, 0-shot) when the Top $K_{\text{mask}}$ regions on the attention map are masked.}
    \label{tab: performance degradation when masked}
    \setlength{\tabcolsep}{3pt} 
    \begin{tabular}{@{}cccccc@{}}
        \toprule
        $K_{\text{mask}}$ & 0 & 10 & 20 & 30 & 50 \\ \midrule 
        M3CoT & 0\% & +0.05\% & -0.11\% & -0.36\% & -0.73\% \\ 
        ScienceQA & 0\% & -0.03\% & -0.08\% & -0.26\% & -0.55\% \\ 
\bottomrule
    \end{tabular}
\end{table}

\subsection{Experiment for Text-vision Granularity Imbalance in Necessity Check}
\label{app:supplementary for granularity imbalance in necessity check in motivation 1}

As mentioned in the necessity check in Section~\ref{sec: motivation for idea 1} and analyzed in Section~\ref{app:semantic_relevance}, in the vast majority of cases (specifically, over 75\%), the VLM fails to distribute its attention toward the truly crucial regions, i.e., those containing salient task-relevant information.

In this section, we present experimental evidence demonstrating that mitigating the \textit{text-vision granularity imbalance} inherent in VQA benchmarks significantly enhances the VLM's ability to accurately focus on salient visual content. Specifically, this is achieved by eliciting the VLM to extract the visual information from the informative image in textual form and transferring it to the otherwise brief raw query.

\paragraph{Experimental Setup.}
We adhere to the experimental setup detailed in Section~\ref{app:semantic_relevance}. The mitigation of granularity imbalance is implemented via CAG component proposed in Section~\ref{sec: method 1}. To assess whether mitigating granularity imbalance significantly improves the VLM's attention distribution, we compare the Intersection over Union (IoU) between the most attended regions and the ground-truth crucial regions before and after the mitigation of imbalance. Furthermore, Wilcoxon signed-rank test is adopted to verify the statistical significance of the improvement.

\paragraph{Experimental Results.}
The results are reported in Table~\ref{tab:imbalance}. Evidently, by enriching the originally brief textual query with additional semantic anchors, the VLM is able to more accurately localize the critical regions within the input image. Moreover, we observe that this improvement becomes increasingly pronounced as the length of the CAG-generated description increases. The Wilcoxon signed-rank test confirms that these improvements are highly statistically significant ($p < 0.001$).

\begin{table}[t]
\centering
\caption{Analysis of the impact of mitigating text-vision granularity imbalance on VLM's attention alignment. \textit{Raw Query} denotes the baseline using the brief textual query, while \textit{Enhanced Query} incorporates the fine-grained image description generated by CAG. }
\label{tab:imbalance}
\resizebox{\linewidth}{!}{
\begin{tabular}{lccc}
\toprule
\textbf{Input Context} & \textbf{Mean IoU (\%)} & \textbf{$\Delta$ IoU} & \textbf{\textit{p}-value} \\
\midrule
Raw Query (Baseline) & 4.5 & - & - \\
\midrule
Enhanced Query (48 Tokens) & 12.2 & +7.7 & $< 0.001$ \\
Enhanced Query (96 Tokens) & {18.6} & {+14.1} & ${< 0.001}$ \\
Enhanced Query (144 Tokens) & {24.1} & {+19.6} & ${< 0.001}$ \\
\bottomrule
\end{tabular}
}
\end{table}

\section{Explanation for Figures}

\subsection{Explanation for Figure \ref{fig: 22nd example - motivation 2}}
\label{app: explanation for fig 2}

\begin{figure}[t]
    \centering
    \begin{subfigure}[t]{0.23\textwidth}
        \centering
        \includegraphics[width=\linewidth]{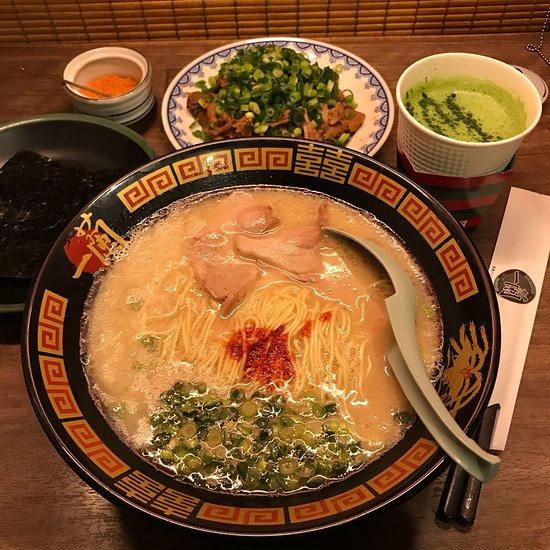}
        \subcaption{Close-up meal photo (query: ``What's the name of the restaurant serving these dishes?'').}
        \label{fig: 22nd example - motivation 1-a}
    \end{subfigure}
    \hfill
    \begin{subfigure}[t]{0.23\textwidth}
        \centering
        \includegraphics[width=\linewidth]{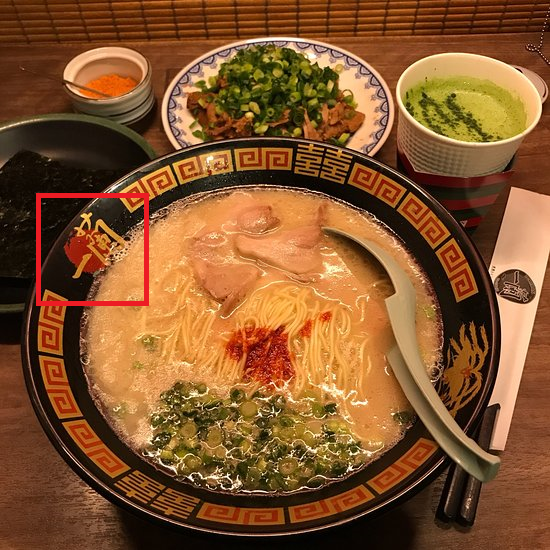}
        \subcaption{Ground-truth cue is localized to a tiny area on the bowl rim (red box).}
        \label{fig: 22nd example - motivation 1-b}
    \end{subfigure}
    \caption{An intuitive example from LLaVA-W shows an ICHIRAN meal close-up. The left image is the raw image, while the small region containing the truly crucial information for answering the question is highlighted with the red box. }
    \label{fig: 22nd example - motivation 1}
\end{figure}

The example adopted in Section~\ref{sec: motivation for idea 2} is the 22nd question in the LLaVA-W benchmark.
The query for this question is: \textit{``What's the name of the restaurant serving these dishes?''} and the associated image is a close-up photo of a meal at ICHIRAN, as shown in Figure~\ref{fig: 22nd example - motivation 1-a}. This is a representative and challenging example because the query is extremely brief, while the image contains highly dense information: although it includes ramen, side dishes, and tableware, the {truly task-relevant} visual evidence occupies only a tiny portion of the image---a small region on the left side of the bowl's rim (Figure~\ref{fig: 22nd example - motivation 1-b}).

Figure~\ref{fig: 22nd example - motivation 2} compares the visual evidence selected by different methods.
The left and right subfigures visualize the top-3 sets of attention-based regions and information-driven regions, respectively. The first, second, and third sets are colored red, purple, and blue, respectively; each set contains $72$ patches.

As shown in Figure~\ref{fig:22-a}, the attention-based regions are scattered, making it difficult to convey complete and concrete visual information. More importantly, they fail to cover the small area that contains the truly crucial evidence. In contrast, as shown in Figure~\ref{fig:22-b}, the information gain-based selection guides the VLM to first focus on the inner rim of the bowl (red) accurately, where the critical information is contained. Although the VLM does not yield the final answer in this region, this indicates a correct line of reasoning, as the ground truth is situated in a highly similar area---a nearby location also on the inner wall of the bowl. Subsequently, the region ranked third (blue) precisely encompasses a large portion of the restaurant's name, which is just the answer to the question. This suggests that even for a challenging case where the text-vision granularity is highly disparate, information gain serves as a better foundation for region selection.

\subsection{Quick Guide to Framework Overview}
\label{app:fig_quick_guide}

Figure~\ref{fig: framework for avp} illustrates AIM-CoT as a Trigger---Select---Insert loop over the VLM's multimodal context window during \textit{Autoregressive Generation} (blue, bottom).

\textbf{CAG (green, left)} performs \textit{Text Context Enhancement} by eliciting a fine-grained, query-conditioned image description and appending it to the original query. The enhanced query leads to \textit{Refined Attention}, which both improves model reasoning and provides more reliable region candidates for subsequent selection.

During decoding, \textbf{DAT (yellow, top)} tracks the \textit{Quantification of Text-to-vision Attention Shift}; when the shift is significant (i.e., $\Delta A_v(t)>\delta$), the \textit{Triggering Condition} is met and a visual insertion is activated.

Once triggered, \textbf{AVP (purple, middle)} conducts \textit{Candidate Region Set Construction} by merging attention-driven and exploratory regions, performs \textit{Information Gain (IG) Quantification} (i.e., uncertainty reduction) for candidates, and runs a greedy \textit{Sequential Selection Process} for $K$ steps. In each step, AVP selects the region with the highest information gain and inserts it into the context window; the next step's selection is conditioned on the updated context.

Finally, the inserted regions are accompanied by a \textit{Critical Integration Strategy} (safety instruction) that treats this visual evidence as supplementary and requires consistency checking in the VLM. Once included in the context window, they help refine subsequent next-token predictions.

\section{Statistical Significance Testing}
\label{app:statistical_test}

To rigorously validate the superiority of AIM-CoT, we conduct statistical significance tests comparing our method against the runner-up performance. The one-sided approach is chosen because our hypothesis specifically posits that AIM-CoT outperforms the baseline.

\paragraph{M3CoT and ScienceQA.}
Since Accuracy (ACC) is derived from binary outcomes, we employ the {one-sided McNemar's Test}. This focuses on assessing whether the number of instances where AIM-CoT corrects ICoT's errors significantly exceeds the number of instances where it introduces new errors.

\paragraph{LLaVA-W.}
For the ROUGE-L metric, which involves continuous scores, we utilize the {one-sided Wilcoxon Signed-Rank Test} (alternative hypothesis: AIM-CoT outperforms the runner-up). This non-parametric paired test evaluates whether the distribution of improvement scores is significantly positive.
 
\vspace{0.5em}
\noindent In Table~\ref{tab: performance_chameleon}, entries marked with $\dagger$ and $\ddagger$ denote statistical significance with $p < 0.05$ and $p < 0.01$, respectively. The detailed p-values and statistics are shown in Table \ref{tab:app_stat_tests}.

\begin{table*}[t]
    \centering
    \caption{Statistical significance tests comparing AIM-CoT against the runner-up under the same backbone/setting.
    Entries report one-sided $p$-values with test statistics in parentheses.}
    \label{tab:app_stat_tests}

    \scriptsize
    \setlength{\tabcolsep}{0pt}
    \renewcommand{\arraystretch}{1.05}

    \begin{minipage}[t]{0.49\textwidth}
        \centering
        \begin{tabular*}{\linewidth}{@{\extracolsep{\fill}}lcc}
            \toprule
            \multirow{2}{*}{Dataset} & \multicolumn{2}{c}{0/1-shot} \\
            \cmidrule(lr){2-3}
            & 0-shot & 1-shot \\
            \midrule

            \multicolumn{3}{c}{\cellcolor{gray!10}\textbf{Chameleon-7B}} \\
            \midrule
            M3CoT (ACC)     & 0.0004$^{\ddagger}$(11.5)  & 0.0228$^{\dagger}$(4) \\
            ScienceQA (ACC) & 0.0000$^{\ddagger}$(22.7)  & 0.0246$^{\dagger}$(3.87) \\
            LLaVA-W (R-L)   & 0.00001$^{\ddagger}$(1502) & 0.00002$^{\ddagger}$(1477) \\
            \midrule

            \multicolumn{3}{c}{\cellcolor{gray!10}\textbf{Janus-Pro-7B}} \\
            \midrule
            M3CoT (ACC)     & 0.0000$^{\ddagger}$(15.6)  & 0.0006$^{\ddagger}$(10.4) \\
            ScienceQA (ACC) & 0.0003$^{\ddagger}$(11.7)  & 0.0000$^{\ddagger}$(23) \\
            LLaVA-W (R-L)   & 0.00012$^{\ddagger}$(1414) & 0.00121$^{\ddagger}$(1327) \\
            \bottomrule
        \end{tabular*}
    \end{minipage}
    \hfill
    \begin{minipage}[t]{0.49\textwidth}
        \centering
        \begin{tabular*}{\linewidth}{@{\extracolsep{\fill}}lcc}
            \toprule
            \multirow{2}{*}{Dataset} & \multicolumn{2}{c}{0/1-shot} \\
            \cmidrule(lr){2-3}
            & 0-shot & 1-shot \\
            \midrule

            \multicolumn{3}{c}{\cellcolor{gray!10}\textbf{Qwen2-VL-7B}} \\
            \midrule
            M3CoT (ACC)     & 0.0224$^{\dagger}$(4.02)  & 0.0224$^{\dagger}$(4.02) \\
            ScienceQA (ACC) & 0.0223$^{\dagger}$(4.03)  & 0.0241$^{\dagger}$(3.91) \\
            LLaVA-W (R-L)   & 0.02035$^{\dagger}$(1193) & 0.03024$^{\dagger}$(1170) \\
            \midrule

            \multicolumn{3}{c}{\cellcolor{gray!10}\textbf{Qwen2.5-VL-32B}} \\
            \midrule
            M3CoT (ACC)     & 0.0075$^{\ddagger}$(5.92) & 0.0012$^{\ddagger}$(9.18) \\
            ScienceQA (ACC) & 0.0028$^{\ddagger}$(7.67) & 0.0107$^{\dagger}$(5.3) \\
            LLaVA-W (R-L)   & 0.00039$^{\ddagger}$(1371)& 0.00000$^{\ddagger}$(1538) \\
            \bottomrule
        \end{tabular*}
    \end{minipage}

    \vspace{2pt}
    \scriptsize
    \raggedright
    \textit{Notes:} $^{\dagger}p<0.05$, $^{\ddagger}p<0.01$ (one-sided). For ACC. (M3CoT/ScienceQA) we use McNemar's test and report the $\chi^2$ statistic in parentheses; for ROUGE-L on LLaVA-W we use the Wilcoxon signed-rank test and report the $W$ statistic in parentheses.
\end{table*}

\section{Validation of Motivation in Section \ref{sec: motivation for idea 3}: Attention Shifts as Dynamic Triggers} \label{app: motivation for idea 3}

\textbf{Experimental Setup.} We take ICoT \cite{icot} as a baseline model, which is required to answer all questions from the LLaVA-W benchmark in a 0-shot setting, with ROUGE-L used as the evaluation metric. The hyper-parameters follow the default settings of the open-source implementation for ICoT, and all experiments are conducted with the Chameleon-7B backbone.

\textbf{Formal Definition of Attention Shifts.} To analyze attention shifts, we examine the averaged attention maps across all attention heads in the last three layers of the VLM during the prediction of each token $t$, following existing research \cite{last_layer_1,last_layer_2,last_layer_3}. The model's total attention scores allocated to the visual and text components of the input are  respectively measured as follows:
\begin{align}
    A_{vision}(t)&=\sum_{i\in \text{indices of $C_{vision}$}} \bar{a}_{t,i}, \\
    A_{text}(t)&=\sum_{j\in \text{indices of $C_{text}$}} \bar{a}_{t,j},
\end{align}
where $C_{vision}, C_{text}$ are the visual and text information within the context, respectively. Then, the shift in attention from the textual to the visual modality while generating token $t$ is defined as follows:
\begin{align}
    \delta_t= A_{vision}(t)-A_{vision}(t-1).
\end{align}
$\Delta_k=[\delta_1,\delta_2,\cdots,\delta_{|\Delta_k|}]$ encompasses the model's attention shifts for each token when answering the arbitrary $k$-th question, where $|\Delta_k|$ is the number of tokens for answering the $k$-th question.

\textbf{Formal Definition of Scores.} For the predictions generated by the baseline model, the ROUGE-L scores are given by $List\_R=[R_1,R_2,\cdots,R_{|List\_R|}]$, where $R_k$ is the score for the model's response to the $k$-th question, and $|List\_R|$ is the number of questions within the benchmark.

Based on these concepts, we design a two-part experiment:

\textbf{Experiment 1: Correlation Analysis.} We investigate the relationship between the proportion of visual insertions under significant attention shifts and the score of the corresponding generated prediction.

First, to identify whether a visual insertion is conducted during a significant attention shift, we define a high attention growth threshold, $\delta_k^{(h)}$ for the $k$-th response ($\delta_k^{(h)}$ is set to the 80\% upper quantile of $\Delta_k$ by default). An insertion is considered to have been conducted under a significant shift and referred to as a \textit{synchronized insertion} if and only if its corresponding attention shift value exceeds the threshold $\delta_k^{(h)}$.

Next, since the model can conduct multiple insertions per response for a question, we calculate $P_k$, the proportion of synchronized insertions out of the total number of insertions for the $k$-th question.

Finally, since the proportions of synchronized insertions $[P_1,P_2,\cdots,P_{|List\_R|}]$ and the ROUGE-L scores for all the questions $[R_1,R_2,\cdots,R_{|List\_R|}]$ are obtained, the Pearson Correlation coefficient can be computed. Specifically, the Pearson Correlation is 0.2166 with a p-value of 0.048, which suggests that the proportions of the synchronized insertions and the corresponding score are significantly positively related to each other.

\textbf{Experiment 2: Group Analysis.} We investigate the relationship between the proportion of synchronized insertions and the quality of the model's response.

To group the generated predictions according to response quality, we establish high- and low-scoring groups. All predictions are ranked in descending order by their ROUGE-L scores. The top 30\% form the high-scoring group (high-quality responses) $G_h$, and the bottom 30\% form the low-scoring group (low-quality responses) $G_l$.

Then, we calculate the mean proportion of synchronized insertions for groups $G_h, G_l$, which are denoted as $\bar{P}_h, \bar{P}_l$, respectively.

Finally, the means of the two groups $\bar{P}_h, \bar{P}_l$ are compared, and a t-test is performed to assess the statistical significance of the difference. Specifically, we find that $\bar{P}_h=0.8889, \bar{P}_l=0.5000$, which suggests that in the high-scoring group, approximately 89\% of insertions are the synchronized insertions with significant attention shift from textual input to visual information; in contrast, in the low-scoring group, only about half of the insertions are synchronized insertions. Besides, the p-value of t-test is as low as $0.0019$, which demonstrates that the result is highly statistically significant.

\section{Hyper-parameter Sensitivity Analysis}

In this section, we conduct a comprehensive sensitivity analysis of hyper-parameters within AIM-CoT. 

Specifically, Appendix \ref{app: sensitivity analysis of delta} investigates the impact of the triggering threshold $\delta$ in the DAT module and introduces an adaptive thresholding strategy. Appendix \ref{app: sensitivity analysis of c, k} examines the influence of the candidate set size $N_C$ and selection count $K$ in the AVP module on computational overhead and deployability.

\subsection{Analysis of $\delta$: Sensitivity and Adaptivity} \label{app: sensitivity analysis of delta}

The hyper-parameter $\delta$ within the DAT module serves as a crucial threshold to trigger the AVP module, which inserts salient visual regions to improve the construction of the multimodal CoT. In this section, we provide a comprehensive analysis of this parameter from two perspectives: (1) a sensitivity analysis to understand its impact on performance and triggering frequency, and (2) an exploration of an adaptive thresholding strategy to demonstrate the framework's robustness and generalizability without per-dataset tuning.

\begin{figure}[htbp]
\centering
  \begin{minipage}{0.4\textwidth}
    \centering
     \includegraphics[width=0.95\textwidth]{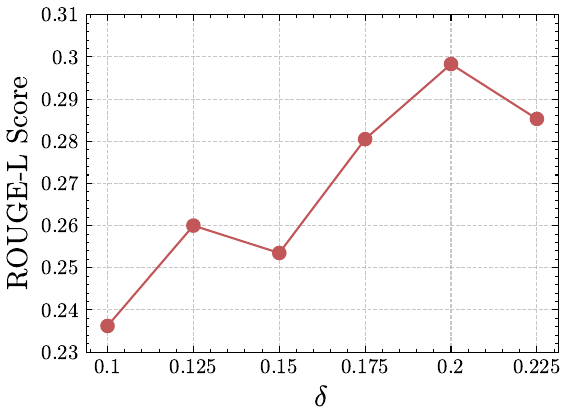}
  \end{minipage}%

  \begin{minipage}{0.4\textwidth}
    \centering
     \includegraphics[width=0.95\textwidth]{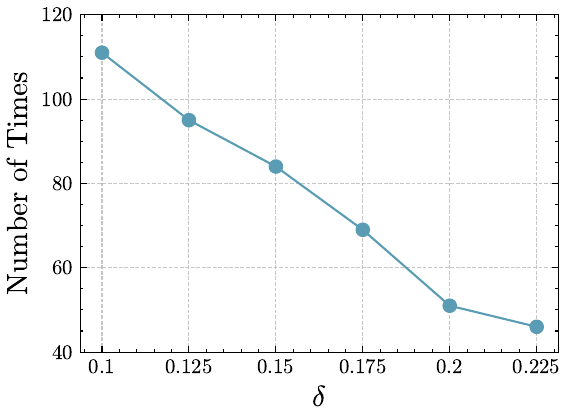}
  \end{minipage}
\caption{Experimental results of the sensitivity analysis of the hyper-parameter $\delta$. The left figure illustrates the performance of AIM-CoT when $\delta$ takes different values, while the right one shows the number of times the AVP module within AIM-CoT is triggered.}
\label{fig: sensitivity analysis of delta}
\end{figure}

\paragraph{Sensitivity Analysis}
We detail a sensitivity analysis of $\delta$ by adjusting it across the range of $[0.1, 0.125, 0.15, 0.175, 0.2, 0.225]$ and examining not only the performance of AIM-CoT but also the number of times the AVP is triggered. The experiments are conducted under the 0-shot setting on the Chameleon-7B backbone and LLaVA-W benchmark. The experimental results are shown in Figure \ref{fig: sensitivity analysis of delta}.

The left figure illustrates that AIM-CoT exhibits limited performance when the threshold, $\delta$, is set too low. This underscores the importance of inserting visual information at critical moments: excessively frequent or inopportune visual insertions can disrupt the VLM's reasoning process, leading to suboptimal performance. As $\delta$ increases, the model's performance progressively improves, reaching its peak at $\delta = 0.2$ (our default setting), which corresponds to a ROUGE-L score of 0.2983. However, a further increase in $\delta$ results in a slight performance degradation. This highlights the criticality of visual information insertion for constructing an interleaved Chain of Thought: an overly stringent threshold excessively impedes the incorporation of visual data, preventing the AVP from supplying the model with necessary visual supplementation in a timely manner.

Conversely, the right figure demonstrates a consistent decrease in the number of times the AVP is triggered as $\delta$ is raised. This showcases the efficacy of $\delta$ as a threshold for modulating the activation frequency of the AVP.

\paragraph{Adaptive Thresholding Strategy}
While our main experiments utilize a fixed $\delta$ tuned for specific benchmarks (e.g., $\delta=0.5$ for M3CoT, and $\delta=0.2$ for ScienceQA and LLaVA-W), we acknowledge that the optimal threshold is intrinsically linked to the granularity of reasoning required by the task. M3CoT involves complex logic with sparse visual queries, leading to sharp, intense attention shifts (requiring a higher threshold to filter noise). In contrast, ScienceQA and LLaVA-W involve generating explanations or descriptions that require continuous visual grounding, resulting in smoother attention shifts (favoring a lower threshold).

To address this variance and enhance the deployability of AIM-CoT without requiring per-dataset tuning, we propose and evaluate an \textbf{Adaptive Z-Score Triggering} mechanism. Instead of a fixed absolute threshold, this method employs a dynamic relative threshold based on the statistical properties of the attention shifts within the current generation context. Specifically, we calculate the Z-score of the current attention shift $\Delta A_{vision}(t)$ relative to the moving average of the previous $W$ tokens:
\begin{equation}
    Z_t = \frac{\Delta A_{vision}(t) - \mu_{t-W:t-1}}{\sigma_{t-W:t-1} + \epsilon},
\end{equation}
where $\mu$ and $\sigma$ are the mean and standard deviation of the attention shifts over a window $W$ (set to 5 by default), and $\epsilon$ is a small constant for numerical stability. The trigger is activated when $Z_t > \lambda$, where $\lambda$ is a universal sensitivity parameter (set to 3).

\begin{table*}[h]
\centering
\caption{Performance comparison between the optimal Fixed Threshold (Tuned) and the Adaptive Z-Score Strategy across three backbones (0-shot setting).}
\label{tab: adaptive threshold}
\resizebox{\textwidth}{!}{%
\begin{tabular}{llcccc}
\toprule
\multirow{2}{*}{Dataset} & \multirow{2}{*}{Method} & \multirow{2}{*}{Setting} & \multicolumn{3}{c}{Performance} \\
\cmidrule(l){4-6}
 &  &  & Chameleon-7B & Janus-Pro-7B & Qwen2.5-VL-32B \\
\midrule
\multirow{2}{*}{M3CoT (ACC.)} & Fixed Threshold & $\delta=0.5$ & 31.4 & 39.7 & 58.7 \\
 & Adaptive Z-Score & $\lambda=3$ & 31.1 & 39.5 & 58.6 \\
\midrule
\multirow{2}{*}{ScienceQA (ACC.)} & Fixed Threshold & $\delta=0.2$ & 53.1 & 56.9 & 76.8 \\
 & Adaptive Z-Score & $\lambda=3$ & 52.9 & 56.7 & 76.8 \\
\midrule
\multirow{2}{*}{LLaVA-W (ROUGE-L)} & Fixed Threshold & $\delta=0.2$ & 29.8 & 35.5 & 46.5 \\
 & Adaptive Z-Score & $\lambda=3$ & 29.5 & 35.4 & 46.3 \\
\bottomrule
\end{tabular}%
}
\end{table*}

As presented in Table \ref{tab: adaptive threshold}, the Adaptive Z-Score strategy demonstrates remarkable robustness. On the powerful Qwen2.5-VL-32B backbone, the adaptive method matches the performance achieved with the tuned threshold on ScienceQA (76.8) and shows negligible deviation on M3CoT (-0.1), suggesting that stronger models exhibit more distinct attention patterns that are easier to capture dynamically. Even on smaller models like Janus-Pro-7B, the performance gap remains minimal (e.g., only 0.1 drop on LLaVA-W). This confirms that while a tuned fixed threshold offers a slight edge, the adaptive strategy provides a highly competitive and deployment-friendly alternative.

\begin{table*}[htbp]
\caption{Unified sensitivity analysis of hyper-parameters $N_C$ and $K$ across 2 benchmarks (M3CoT and LLaVA-W) on 2 VLM backbones (Chameleon-7B and Qwen2.5-VL-32B).
The table reports AIM-CoT's average processing time (seconds) per instance.
Entries where $N_C < K$ are marked with ``-'' as $N_C \geq K$ holds by definition.
}
\label{tab: analysis of c, k (merged)}
\centering

\normalsize

\setlength{\tabcolsep}{10pt} 

\begin{tabular}{@{}c cccc c cccc@{}}
\toprule
\multirow{2}{*}{\textbf{$N_C$}} & \multicolumn{4}{c}{\textbf{M3CoT}} & & \multicolumn{4}{c}{\textbf{LLaVA-W}} \\ 
\cmidrule(lr){2-5} \cmidrule(l){7-10}
 & K=1 & K=3 & K=5 & K=7 & & K=1 & K=3 & K=5 & K=7 \\ \midrule
 
\multicolumn{10}{c}{\textit{\textbf{Backbone: Chameleon-7B}}} \\ \midrule
1 & 12.78 & - & - & - & & 10.61 & - & - & - \\
2 & 12.80 & - & - & - & & 10.70 & - & - & - \\
3 & 12.83 & 13.18 & - & - & & 10.75 & 11.21 & - & - \\
4 & 12.87 & 13.23 & - & - & & 10.84 & 11.36 & - & - \\
5 & 12.90 & 13.30 & 13.41 & - & & 10.92 & 11.51 & 11.79 & - \\
6 & 12.93 & 13.37 & 13.49 & - & & 11.00 & 11.65 & 11.96 & - \\
7 & 12.97 & 13.42 & 13.56 & 13.66 & & 11.08 & 11.80 & 12.14 & 12.38 \\ \midrule

\multicolumn{10}{c}{\textit{\textbf{Backbone: Qwen2.5-VL-32B}}} \\ \midrule
1 & 52.12 & - & - & - & & 42.45 & - & - & - \\
2 & 52.31 & - & - & - & & 42.68 & - & - & - \\
3 & 52.48 & 55.08 & - & - & & 42.91 & 45.75 & - & - \\
4 & 52.75 & 55.45 & - & - & & 43.15 & 46.12 & - & - \\
5 & 52.94 & 55.72 & 56.95 & - & & 43.52 & 46.48 & 47.95 & - \\
6 & 53.18 & 55.98 & 57.34 & - & & 43.88 & 46.85 & 48.44 & - \\
7 & 53.42 & 56.23 & 57.68 & 58.71 & & 44.15 & 47.27 & 48.95 & 50.10 \\ \bottomrule
\end{tabular}

\end{table*}


\subsection{Sensitivity Analysis of $K, N_C$} \label{app: sensitivity analysis of c, k}

Our proposed AIM-CoT incorporates the design of AVP module. In contrast to existing research, AIM-CoT, benefiting from the AVP module, does not simply select the top-$K$ regions with the highest attention scores from the attention map. Instead, it meticulously selects $K$ regions from a total set $C$ of $N_C$ candidate regions to construct the multimodal CoT ($N_C=N+M$). However, this approach may inevitably raise concerns regarding the deployability of AIM-CoT, particularly as the hyper-parameters $N_C$ and $K$ increase.

To investigate this, we conduct experiments to explore the average processing time per instance for AIM-CoT with larger values of $N_C$ and $K$. For the experimental setup, we implement AIM-CoT with both Chameleon-7B and Qwen2.5-VL-32B backbones on M3CoT and LLaVA-W datasets under different combinations of $N_C$ and $K$. The results are presented in Table \ref{tab: analysis of c, k (merged)}. From the results, the following two key insights can be derived:
\begin{itemize}[topsep=0.5mm, partopsep=0pt, itemsep=0pt, leftmargin=10pt]
    \item \textbf{Insensitivity to the growth of the candidate set size $N_C$}: observing any row with a fixed $K$ (e.g., $K=3$ on M3CoT with Chameleon-7B), as $N_C$ increases from 3 to 7, the total processing time rises from 13.18s to 13.42s, a marginal increase of only 0.24s. This implies that each additional candidate region introduces an average overhead of less than 0.06s. This strongly demonstrates that the performance of the AVP module does not degrade sharply with a moderate expansion of the candidate pool, indicating excellent scalability.

    \item \textbf{Diminishing marginal cost with the increase in the number of selections $K$}: considering a fixed column for $N_C$ (e.g., $N_C=7$ on LLaVA-W with Chameleon-7B), as $K$ increases from $1 \rightarrow 3$, $3 \rightarrow 5$, and $5 \rightarrow 7$, the processing time increases by 0.72s, 0.34s, and 0.24s, respectively, which shows a notable decrease in incremental cost. This suggests that the primary computational overhead of the AVP algorithm lies in the initiation of the iterative search (the jump from $K=1$ to $K=3$). Once the iteration begins, the cost of subsequent selection steps is remarkably low, benefiting from efficient mechanisms such as the batch processing and KV Cache.
\end{itemize}

Based on these key insights, our proposed AVP module demonstrates high computational efficiency and robustness when faced with increased computational complexity (i.e., larger $N_C$ and $K$ values).


\begin{figure*}[htbp]
    \centering
    \includegraphics[width=\textwidth]{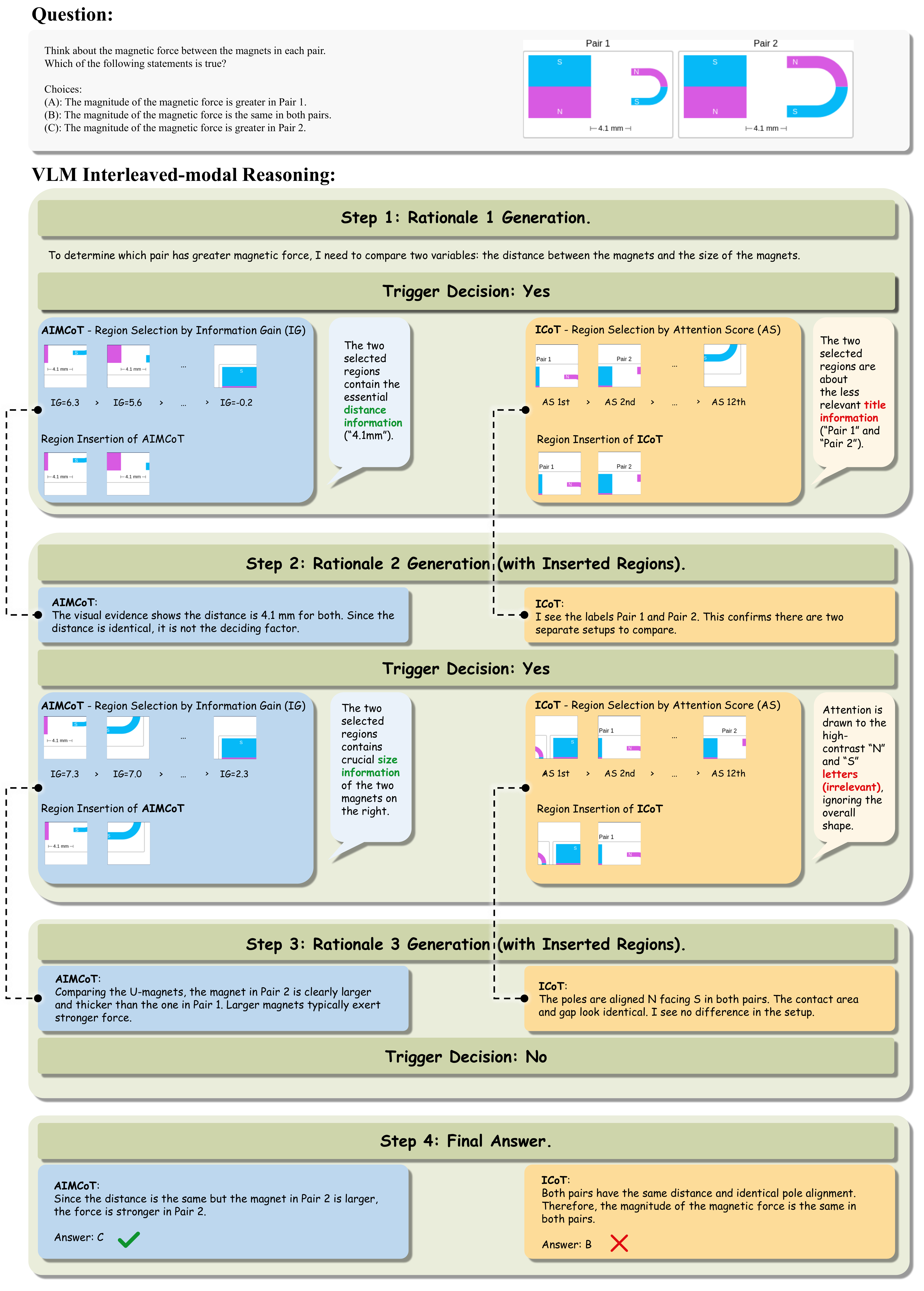}
    \caption{
        \textbf{Qualitative comparison on ScienceQA (Question 244) with Chameleon-7B.} 
        \textbf{Left (Blue):} AIM-CoT actively selects regions based on Information Gain (IG). It correctly focuses on the \textit{distance} (Step 1) and \textit{magnet size} (Step 2), which are the actual visual variables required to solve the physics question.
        \textbf{Right (Yellow):} The baseline ICoT, relying on raw Attention Scores (AS), fails to filter out noise. It attends to irrelevant text (``Pair 1'') and decorative letters (``N/S'') rather than the physical attributes of the objects, resulting in reasoning failure.
    }
    \label{fig:scienceqa_magnet}
\end{figure*}

\section{Qualitative Case Study: Information Gain vs. Attention}
\label{app:qualitative_case_IG_vs_AS}

To complement the quantitative analyses in Sections~\ref{app:semantic_relevance} and~\ref{app:human_alignment_gpt4v}, we provide a step-by-step visualization of the reasoning process to demonstrate the efficacy of our proposed AVP module.

Figure~\ref{fig:scienceqa_magnet} presents a qualitative comparison between AIM-CoT (Ours) and the baseline ICoT on a challenging question from ScienceQA (Question ID: 244) using the Chameleon-7B \cite{chameleon} backbone. 
As observed, the {Information Gain-driven} selection in AIM-CoT successfully identifies task-relevant visual evidence (e.g., \textit{magnet size} and \textit{distance}) that directly contributes to the correct reasoning chain. 
In contrast, the {Attention-driven} baseline is distracted by high-frequency but irrelevant features, such as the text labels (``Pair 1'') or high-contrast letters (``N'', ``S''), leading to hallucinated reasoning and an incorrect answer. This visual evidence reinforces our hypothesis that high-attention regions do not necessarily equate to high-value information for reasoning.

\section{Comprehensive Analysis of AVP Component}

In this appendix, we conduct a comprehensive analysis of the AVP component to validate its design choices and theoretical foundations. The detailed analyses are organized as follows:
\begin{itemize}[topsep=0.5mm, partopsep=0pt, itemsep=0pt, leftmargin=10pt]
    \item Appendix \ref{app: ablation study on C construction} compares different strategies for constructing the candidate set to identify the most effective composition.
    \item {Appendix \ref{app: proportion of sources}} quantifies the specific contribution of the exploratory set by analyzing the source distribution of the selected regions.
    \item {Appendix \ref{app:avp_robustness}} evaluates the robustness of AVP against adversarial attention noise, demonstrating its intrinsic noise-rejection capability.
    \item {Appendix \ref{app: proof for function F}} provides a theoretical justification for (1) the approximate submodularity of the information gain function and (2) the optimality of our greedy selection algorithm.
\end{itemize}

\subsection{Construction of Candidate Set $C$: Method Comparison}
\label{app: ablation study on C construction}

In this section, we investigate the influence of different compositions of the total set $C$ on the performance of our proposed AIM-CoT. We specifically examine two primary configurations: 

\textbf{Constructing $C$ using only $C_{attn}$ or $C_{exp}$.} 
For the latter, we evaluate three distinct construction methodologies for $C_{exp}$: (a) $C_{rand}$: uniform random sampling; (b) $C_{ss}$: the selective search algorithm \cite{ss}, which is the seminal region proposal method utilized in R-CNN \cite{rcnn}; (c) $C_{fsam}$: FastSAM \cite{fsam}, a computationally efficient variant of the foundational vision segmentation model, SAM \cite{sam}.

\textbf{Constructing $C$ using both $C_{attn}$ and $C_{exp}$.} 
Similarly, as for $C_{exp}$, we also consider its diversified construction, including $C_{rand}, C_{ss}$, and $C_{fsam}$. 

It is worth noting that although the construction is diverse, the size of $C$ remains consistent. When $C$ is composed of $C_{attn}$ and $C_{exp}$, the two each account for half. The experimental results are shown in Table \ref{tab: exploratory set comparison}.

As observed, the combination of the two sets (i.e., $C=C_{attn} \cup C_{exp}$) invariably yields superior performance for AIM-CoT compared to configurations where either $C=C_{attn}$ or $C=C_{exp}$ is used exclusively. This highlights the importance of diversifying the sources of candidate visual regions.

When comparing the different construction methods for $C_{exp}$, the performance gap among models is marginal when used in conjunction with $C_{attn}$. Specifically, despite its simplicity, random sampling achieves highly competitive results, which motivates our choice to adopt it as the default method for constructing $C_{exp}$. Intuitively, the advantage of random sampling lies in its ability to provide regions across different parts of the image unbiasedly with maximal spatial diversity.

\begin{table*}[t]
\centering
\caption{Performance comparison of AIM-CoT variants on the basis of different constructions of the candidate set $C$.}
\label{tab: exploratory set comparison}
\begin{tabular}{@{}c|c|c@{}}
\toprule
Construction of $C$ & M3CoT (ACC.) & LLaVA-W (ROUGE-L)\\ \midrule
$C_{attn}\cup C_{rand} ~~~(C_{exp}=C_{rand})$ & 31.4 & 29.8  \\
$C_{attn}\cup C_{ss}~~~(C_{exp}=C_{ss})$ & 31.2 & 29.5 \\
$C_{attn}\cup C_{fsam}~~~(C_{exp}=C_{fsam})$ & 31.0 & 29.6 \\
\midrule
$C_{attn}$ & 30.8 & 28.9 \\ 
$C_{rand}$ & 30.4 & 28.6 \\ 
$C_{ss}$ & 30.3 & 28.7 \\ 
$C_{fsam}$ & 29.9 & 27.7 \\ 
\bottomrule
\end{tabular}
\end{table*}

\subsection{Source Distribution Analysis: Quantifying Contribution of Exploratory Set $C_{exp}$}\label{app: proportion of sources}

In this section, we examine the distribution of sources for the visual regions selected by the AVP module of AIM-CoT. These regions are drawn from two sets, $C_{attn}$ and $C_{exp}$, with their respective selection proportions denoted as $P_{attn}$ and $P_{exp}$. Intuitively, $P_{exp}$ reflects the significance of incorporating the exploratory set $C_{exp}$ to construct a better multimodal CoT. A larger value of $P_{exp}$ indicates that the exploratory set $C_{exp}$ makes a greater contribution by providing informative salient regions to AIM-CoT, and vice versa.

\textbf{Experimental Setup}
The experiments are conducted on the M3CoT and LLaVA-W benchmarks. Our proposed AIM-CoT is implemented with the Chameleon-7B backbone under a default 0-shot setting. To ensure the reliability of the results, we repeat each experiment three times on both benchmarks.

\textbf{Results and Analysis}
As presented in Table~\ref{tab:source_distribution}, although the value of $P_{exp}$ fluctuates across different experimental runs on the same benchmark, it remains consistently around 20\% on M3CoT and 30\% on LLaVA-W. This indicates that the influence of stochastic factors on the source distribution of the selected regions is limited, which validates our rationale of using this metric as a reflection of the relative importance of $C_{attn}$ and $C_{exp}$. Furthermore, we observe that $P_{exp}$ is significantly greater than zero. This demonstrates that the exploratory set $C_{exp}$ consistently serves as a critical component of the total candidate set $C$, contributing a substantial portion of the informative regions for AIM-CoT.

\begin{table}[htbp]
\centering
\caption{Proportion of salient regions selected by the AVP module of our proposed AIM-CoT from the exploratory set $C_{exp}$.}
\label{tab:source_distribution}
\begin{tabular}{@{}cccc@{}}
\toprule
Experiment Number & 1 & 2 & 3 \\ \midrule
M3CoT & 17.25\% & 20.44\% & 27.27\% \\
LLaVA-W & 31.33\% & 25.77\% & 26.67\% \\ \bottomrule
\end{tabular}
\end{table}

\subsection{Robustness of AVP Against Attention Noise}
\label{app:avp_robustness}

In Section~\ref{sec: method 2}, we posit that AVP possesses an intrinsic noise-rejection capability, formalized in Equation~\ref{eq:ig}, allowing it to filter out high-attention regions that yield negligible information gain. The ability to inherently identify and reject noisy or corrupted signals without supervision is a highly desirable property for robust machine learning systems~\cite{pyliu2,pyliu3}. To empirically validate this resilience against error propagation (e.g., when upstream modules produce misleading attention maps), we conduct an adversarial noise injection experiment.

\textbf{Experimental Setup.} We design an \textbf{Adversarial Noise Injection} protocol. For each image, we randomly select an irrelevant background patch (verified to have no overlap with ground-truth objects) and artificially override its attention score to be the global maximum in the attention map. We then compare the region selection behavior and downstream performance of two strategies using this corrupted attention map:
(1) \textbf{Top-K Selection (Baseline):} Selects regions strictly based on attention scores.
(2) \textbf{AVP (Ours):} Selects regions based on Information Gain (IG).
Consistent with Appendix~\ref{app:cag_reliability}, we evaluate on M3CoT, ScienceQA, and LLaVA-W using Chameleon-7B and Qwen2-VL-7B backbones.

\textbf{Evaluation Metrics.}
(1) \textbf{Noise Rejection Rate (NRR):} The percentage of instances where the model successfully avoids selecting the injected noise patch, despite it having the highest attention score.
(2) \textbf{Performance Drop ($\Delta$):} The decline in task performance compared to the clean (non-adversarial) setting. Smaller $\Delta$ indicates higher robustness.

\begin{table*}[h]
\centering
\caption{Robustness analysis against adversarial attention noise. NRR denotes Noise Rejection Rate ($\uparrow$ higher is better); $\Delta$ denotes performance drop ($\downarrow$ lower is better, closer to 0).}
\label{tab:avp_robustness}
\resizebox{\textwidth}{!}{%
\begin{tabular}{llcccccc}
\toprule
\multirow{2}{*}{Backbone} & \multirow{2}{*}{Method} & \multicolumn{2}{c}{M3CoT} & \multicolumn{2}{c}{ScienceQA} & \multicolumn{2}{c}{LLaVA-W} \\
\cmidrule(lr){3-4} \cmidrule(lr){5-6} \cmidrule(lr){7-8}
 &  & NRR ($\%$) $\uparrow$ & $\Delta$ Acc. $\downarrow$ & NRR ($\%$) $\uparrow$ & $\Delta$ Acc. $\downarrow$ & NRR ($\%$) $\uparrow$ & $\Delta$ ROUGE-L $\downarrow$ \\
\midrule
\multirow{2}{*}{Chameleon-7B} & Top-K & 0.0 & -4.2 & 0.0 & -3.8 & 0.0 & -5.1 \\
 & \textbf{AVP (Ours)} & \textbf{96.4} & \textbf{-0.2} & \textbf{98.1} & \textbf{-0.1} & \textbf{95.5} & \textbf{-0.4} \\
\midrule
\multirow{2}{*}{Qwen2-VL-7B} & Top-K & 0.0 & -3.5 & 0.0 & -2.9 & 0.0 & -4.4 \\
 & \textbf{AVP (Ours)} & \textbf{97.8} & \textbf{-0.1} & \textbf{99.2} & \textbf{0.0} & \textbf{96.7} & \textbf{-0.2} \\
\bottomrule
\end{tabular}%
}
\end{table*}

As shown in Table~\ref{tab:avp_robustness}, the contrast between the two methods is stark:

First, {Top-K is fundamentally vulnerable to attention errors.} By definition, Top-K has an NRR of 0.0\%, as it blindly trusts the manipulated attention scores. Consequently, it forces the inclusion of irrelevant visual noise, leading to significant performance degradation across all benchmarks (e.g., dropping 4.2\% accuracy on M3CoT with Chameleon-7B).

Second, {AVP demonstrates exceptional noise immunity.} Even when the noise patch is forced to have the highest attention priority, AVP successfully rejects it in over 95\% of cases (NRR $>95\%$). This confirms that AVP correctly identifies that the noise patch, despite its high attention score, provides minimal reduction in uncertainty (low IG). As a result, the performance drop is negligible ($\Delta \approx 0$), proving that AVP effectively breaks the chain of error propagation, ensuring robust reasoning even when the guidance from the attention map is unreliable.

\subsection{Theoretical Justification: Approximate Submodularity and Optimality of Greedy Selection} \label{app: proof for function F}

To offer a more comprehensive insight into the motivation for employing a greedy algorithm, this section provides a thorough analysis. We emphasize that for functions that are not theoretically submodular, a greedy approach remains one of the conventional methods for addressing their maximization, as established in recognized works \cite{greedy2,greedy3,greedy4,greedy5,greedy6}. In this part, we conduct meticulously designed experiments to investigate the extent to which the information gain function $F$ approximates submodularity. The experimental results reveal that $F$ empirically exhibits significant submodular characteristics. This finding motivates us to follow established works \cite{greedy2,greedy3,greedy4,greedy5,greedy6} to propose a greedy algorithm to solve the maximization problem for $F$. The analysis is detailed as follows:

Firstly, we would like to introduce the definition of a submodular function. According to existing research \cite{nphard1&greedy1}, a function $f$ is a submodular function if it satisfies 
\begin{align}\label{ineq: definition of sub}
    f(A\cup \{R_i\})-f(A)\geq f(B\cup \{R_i\})-f(B) 
\end{align}
for any sets $A\subseteq B\subset C$ and any element that satisfies $R_i \in C\backslash B$. In our scenario, the Inequality \ref{ineq: definition of sub} is written as 
\begin{align}
    F(A\cup \{R_i\})-F(A)\geq F(B\cup \{R_i\})-F(B) 
\end{align}
for any $A\subset B\subset C$ and any $R_i \in C\backslash B$, which means that the information gain from incorporating a visual region exhibits a diminishing returns property.

To demonstrate this empirically, we design the experiment detailed as follows, aiming to show that for two sets of regions of different sizes, $S_{small} \subset S_{large} \subset C$, the information gain from incorporating a given visual region $R_{test} \in C \setminus S_{large}$ into the context of a VLM is greater when $R_{test}$ is added to $S_{small}$ than when it is added to $S_{large}$, ceteris paribus.

\textbf{Experimental Setup.}
In our experimental design, each time the AVP process is triggered to select salient regions, we first execute it to select $K_{small}$ regions from the total candidate pool $C$ to form the set $S_{small}$. Subsequently, building upon $S_{small}$, we select an additional $K_{large}-K_{small}$ regions to construct the set $S_{large}$, where $K_{small}$ and $K_{large}$ are the respective set sizes. This construction inherently ensures that $S_{small} \subset S_{large}$.

Next, to compute the information gain contributed by a given region, we randomly sample a region $R_{test}$ from $C \setminus S_{large}$. We then calculate the VLM's information contents, which are denoted as $U_{s}, U_{s}^*, U_{l}, \text{and } U_{l}^*$, when the context incorporates (1) $S_{small}$, (2) $S_{small}\cup \{R_{test}\}$, (3) $S_{large}$, and (4) $S_{large}\cup \{R_{test}\}$, respectively. We expect to observe in the majority of cases that:
\begin{align}
U_{s}^* - U_{s} \geq U_{l}^* - U_{l}.
\end{align}

We conduct experiments on the M3CoT and LLaVA-W benchmarks, setting $K_{small} \in \{2,3,4,5\}$ and $K_{large} = K_{small}+1$ for simplicity. In terms of evaluation, we record the proportion of instances for which the inequality $U_{s}^* - U_{s} \geq U_{l}^* - U_{l}$ holds, and further introduce a Binomial Test to rigorously examine the significance of the results.

\textbf{Experimental Results.}
The experimental results are presented in Table \ref{tab: empirical proof for submodularity}. As we can see, the Inequality \ref{ineq: definition of sub} holds in most instances across all settings and datasets. Furthermore, to confirm the significance of the obtained results, we introduce the Binomial Test, an exact statistical procedure for assessing the extent to which experimental outcomes with a binary structure are attributable to chance alone. The p-values, presented in Table \ref{tab: empirical proof for submodularity}, are all substantially below the 0.05 significance level. This demonstrates that the information gain function $F$ behaves in a manner that is empirically near-submodular, which motivates us to follow existing research \cite{greedy2,greedy3,greedy4,greedy5,greedy6} where greedy algorithms are proposed to solve the problem of maximizing approximately submodular functions.

\begin{table}[htbp]
\caption{Proportions of instances on M3CoT and LLaVA-W benchmarks for which the approximate submodularity of information gain function $F$ is manifested. The backbone is Chameleon-7B and the model is our proposed AIM-CoT. $K_{large}$ is set to $K_{small}+1$ for simplicity. The significance levels of these results are listed below them.}
\label{tab: empirical proof for submodularity}
\centering
\resizebox{\columnwidth}{!}{
\begin{tabular}{@{}ccccc@{}}
\toprule
$K_{small}$ & 2 & 3 & 4 & 5 \\ \midrule
M3CoT (n=2318) & 72.00\% & 62.99\% & 67.04\% & 61.09\% \\
p-value & <1e-6 & <1e-6 & <1e-6 & <1e-6 \\
\midrule
LLaVA-W (n=60) & 61.67\% & 68.33\% & 61.67\% & 63.33\% \\ 
p-value & 0.0462 & 0.0031 & 0.0462 & 0.0249 \\
\bottomrule
\end{tabular}
}
\end{table}

\section{Extensive Ablation Studies on More Backbones}
\label{app:extensive ablation studies}

Section~\ref{sec: ablation study (ours)} presents an ablation study on the Chameleon-7B backbone. The results provide two key insights as follows: 
\begin{itemize}[topsep=0.5mm, partopsep=0pt, itemsep=0pt, leftmargin=10pt]
\item  Each component (CAG, AVP, DAT) contributes substantially to the overall performance. 
\item Even without CAG, combining AVP and DAT is sufficient for AIM-CoT to outperform strong baseline models by a clear margin. This finding highlights the importance of addressing the “what to see” and “when to see it” questions. 
\end{itemize}
To test whether these insights generalize beyond Chameleon-7B, in this section, we extend the same ablation study to two additional VLM backbones: Janus-Pro-7B and Qwen2.5-VL-32B.

As shown in Tables~\ref{tab: ablation_janus} and~\ref{tab: ablation_qwen}, the ablation trends are consistent with those in Section~\ref{sec: ablation study (ours)}: removing any component leads to performance degradation across datasets and backbones, demonstrating the efficacy of our proposed components.

More importantly, the results highlight the central thesis of this study. \textbf{Even without CAG, the combination of AVP and DAT already yields strong performance and is sufficient to surpass strong baselines, including ICoT}. This demonstrates that explicitly and effectively addressing \textit{what to see} (AVP) and \textit{when to see it} (DAT) is the key driver of AIM-CoT's advantage.

Meanwhile, {CAG remains a non-trivial contributor}: although its absolute gains are comparatively smaller, it consistently improves results across all settings, suggesting that mitigating the text-vision granularity mismatch further complements AVP+DAT.

\begin{table}[htbp]
    \centering
    \caption{Ablation study of AIM-CoT conducted on Janus-Pro-7B under 0-shot setting.}
    \label{tab: ablation_janus}
    \resizebox{\columnwidth}{!}{%
        \begin{tabular}{@{}lcccc@{}}
            \toprule
            Dataset & AIM-CoT & w/o CAG & w/o AVP & w/o DAT \\
            \midrule
            M3CoT (ACC.) & 39.7 & 39.4 (\textcolor{red}{-0.3}) & 38.6 (\textcolor{red}{-1.1}) & 39.0 (\textcolor{red}{-0.7}) \\
            ScienceQA (ACC.) & 56.9 & 56.7 (\textcolor{red}{-0.2}) & 55.8 (\textcolor{red}{-1.1}) & 56.3 (\textcolor{red}{-0.6}) \\
            LLaVA-W (ROUGE-L) & 35.5 & 34.7 (\textcolor{red}{-0.8}) & 33.5 (\textcolor{red}{-2.0}) & 34.5 (\textcolor{red}{-1.0}) \\
            \bottomrule
        \end{tabular}%
    }
\end{table}

\begin{table}[htbp]
    \centering
    \caption{Ablation study of AIM-CoT conducted on Qwen2.5-VL-32B under 0-shot setting.}
    \label{tab: ablation_qwen}
    \resizebox{\columnwidth}{!}{%
        \begin{tabular}{@{}lcccc@{}}
            \toprule
            Dataset & AIM-CoT & w/o CAG & w/o AVP & w/o DAT \\
            \midrule
            M3CoT (ACC.) & 58.7 & 58.4 (\textcolor{red}{-0.3}) & 57.6 (\textcolor{red}{-1.1}) & 58.1 (\textcolor{red}{-0.6}) \\
            ScienceQA (ACC.) & 76.8 & 76.6 (\textcolor{red}{-0.2}) & 75.9 (\textcolor{red}{-0.9}) & 76.3 (\textcolor{red}{-0.5}) \\
            LLaVA-W (ROUGE-L) & 46.5 & 46.1 (\textcolor{red}{-0.4}) & 44.2 (\textcolor{red}{-2.3}) & 45.0 (\textcolor{red}{-1.5}) \\
            \bottomrule
        \end{tabular}%
    }
\end{table}

\section{In-depth Analysis of Performance Gain Variations Across Backbones}
\label{sec:performance_variance_analysis}

In Table \ref{tab: performance_chameleon}, we observe that AIM-CoT yields varying degrees of improvement across different base models. Specifically, the gains on Chameleon-7B are substantially larger than those on Qwen2-VL-7B, while Qwen2.5-VL-32B exhibits higher improvements than its 7B counterpart. In this section, we analyze the underlying factors contributing to these phenomena from two perspectives: visual encoding mechanisms and model capability scaling.

\subsection{Cross-Family Analysis: Architecture and Visual Encoding}
The distinct performance gap between Chameleon-7B (up to 18.25\% gain on LLaVA-W) and Qwen2-VL-7B (up to 6.14\% gain) can be attributed to their fundamental differences in visual processing:

\begin{itemize}[topsep=0.5mm, partopsep=0pt, itemsep=0pt, leftmargin=10pt]
    \item {Chameleon-7B (Early-Fusion \& Fixed Resolution):} As an early-fusion model utilizing a fixed tokenization strategy, Chameleon-7B often struggles with fine-grained visual details in high-resolution images, leading to the ``myopic'' attention behavior discussed in Section~\ref{sec: motivation for idea 1}. AIM-CoT acts as a \textbf{perceptual correction} mechanism here. By actively cropping and zooming in on salient regions via the AVP module, AIM-CoT directly compensates for the model's native resolution limitations. The improvement is transformative as it solves a ``can't see'' problem, resulting in substantial gains.
    
    \item {Qwen2-VL Series (NaViT \& Dynamic Resolution):} These models employ the NaViT architecture with dynamic resolution support, allowing them to process arbitrary aspect ratios and resolutions natively. Since Qwen2-VL can already perceive visual details relatively clearly, the marginal utility of ``zooming in'' is lower compared to Chameleon. For Qwen models, AIM-CoT serves more as an \textbf{attention optimization} tool rather than a vision repair tool, leading to more moderate base improvements.
\end{itemize}

\subsection{Intra-Family Analysis: Capability Dependence and Scaling Laws}
Within the Qwen family, we observe that the larger Qwen2.5-VL-32B benefits more from AIM-CoT (up to 9.84\% gain) than the smaller Qwen2-VL-7B (up to 6.14\% gain). This counter-intuitive trend (where a stronger baseline yields larger relative gains) highlights the \textbf{capability-dependent nature} of our framework:

\begin{itemize}[topsep=0.5mm, partopsep=0pt, itemsep=0pt, leftmargin=10pt]
    \item {Quality of CAG Generation:} AIM-CoT is a closed-loop system where the Context-enhanced Attention-map Generation (CAG) module relies on the model's own capability to describe the image. The 32B model generates more precise, hallucination-free descriptions than the 7B model. This high-quality context leads to more accurate attention maps, enabling the AVP module to select significantly more valuable image regions.
    
    \item {Interleaved Reasoning Capability:} The efficacy of AIM-CoT also depends on how well the model can leverage the inserted visual patches. Larger models (32B) possess stronger long-context understanding and multi-step reasoning capabilities. They can effectively synthesize the fragmented information provided by the inserted patches to deduce the correct answer, whereas smaller models might struggle to integrate these additional visual cues coherently.
\end{itemize}

In summary, AIM-CoT functions as a mechanism for {Perceptual Correction} on weaker architectures (Chameleon) and {Reasoning Augmentation} on stronger ones (Qwen 32B), demonstrating its adaptability across different model paradigms.

\section{Safety Mechanisms in AIM-CoT}

To ensure the reliability of the reasoning chain and prevent error propagation, AIM-CoT incorporates explicit safety mechanisms at two critical stages.
First, regarding \textbf{Source Quality (Input)}, we apply \textit{Negative Constraints} within the CAG module to strictly ground the generated context and mitigate hallucinations.
Second, regarding \textbf{Integration Safety (Process)}, we employ \textit{Safety Instructions} during the visual insertion, guiding the model to verify relevance and filter out potential noise.
In this section, we analyze the individual contributions of these safeguards to the framework's overall robustness.

\subsection{
Effectiveness of Negative Constraints in CAG Generation
}
\label{app:cag_reliability}

In Section~\ref{sec: method 1}, we introduce negative constraints within the CAG prompt $\mathcal{P}_{CAG}$ to ensure the utmost precision of the generated descriptions. To rigorously assess the necessity of this design and the potential risk of error propagation, we conduct an ablation study focusing on the hallucination rate of the generated descriptions $\mathcal{D}_{CAG}$ and their impact on downstream performance.

\textbf{Experimental Setup.} We contrast the standard AIM-CoT against \textbf{AIM-CoT w/o Constraints}, a variant where the cautionary instructions are removed from $\mathcal{P}_{CAG}$. To demonstrate the generalizability of our design, we implement the framework on two representative backbones: {Chameleon-7B} and {Qwen2-VL-7B}. The experiments are conducted on M3CoT, ScienceQA, and LLaVA-W benchmarks under the 0-shot setting.

\textbf{Evaluation Metrics.}
Two metrics are incorporated for evaluation:
(1) \textbf{Hallucination Rate (HR):} We employ GPT-4v as an external evaluator to identify factual inconsistencies between the image content and the generated description $\mathcal{D}_{CAG}$. HR is reported as the percentage of descriptions containing details unsupported by the visual input.
(2) \textbf{Downstream Performance:} We report Accuracy (ACC.) for M3CoT and ScienceQA, and ROUGE-L for LLaVA-W to measure the final reasoning outcome.

\begin{table*}[h]
\centering
\caption{Ablation study on the effectiveness of Negative Constraints across two backbones. HR denotes Hallucination Rate ($\downarrow$ lower is better); Perf. denotes downstream performance ($\uparrow$ higher is better).}
\label{tab:cag_ablation}
\resizebox{\textwidth}{!}{%
\begin{tabular}{llcccccc}
\toprule
\multirow{2}{*}{Backbone} & \multirow{2}{*}{Method} & \multicolumn{2}{c}{M3CoT} & \multicolumn{2}{c}{ScienceQA} & \multicolumn{2}{c}{LLaVA-W} \\
\cmidrule(lr){3-4} \cmidrule(lr){5-6} \cmidrule(lr){7-8}
 &  & HR ($\%$) $\downarrow$ & Acc. $\uparrow$ & HR ($\%$) $\downarrow$ & Acc. $\uparrow$ & HR ($\%$) $\downarrow$ & ROUGE-L $\uparrow$ \\
\midrule
\multirow{2}{*}{Chameleon-7B} & w/o Constraints & 5.8 & 30.1 & 4.5 & 52.5 & 8.4 & 27.9 \\
 & \textbf{AIM-CoT (Ours)} & \textbf{1.9} & \textbf{31.4} & \textbf{1.2} & \textbf{53.1} & \textbf{2.7} & \textbf{29.8} \\
\midrule
\multirow{2}{*}{Qwen2-VL-7B} & w/o Constraints & 3.5 & 43.9 & 2.8 & 56.9 & 5.1 & 35.1 \\
 & \textbf{AIM-CoT (Ours)} & \textbf{0.8} & \textbf{44.7} & \textbf{0.5} & \textbf{57.4} & \textbf{1.5} & \textbf{36.3} \\
\bottomrule
\end{tabular}%
}
\end{table*}

The results are summarized in Table~\ref{tab:cag_ablation}, from which two key observations are derived:

First, {the CAG module exhibits high intrinsic reliability across backbones.} Even without explicit negative constraints, the hallucination rates remain relatively low (e.g., $<6\%$ on M3CoT for both models). This suggests that the foundational strategy of using VLM-generated descriptions as context is robust and not inherently prone to generating misleading information, regardless of the underlying model architecture.

Second, {negative constraints provide a critical layer of safety.} Despite the solid baseline, the introduction of negative constraints consistently suppresses residual hallucinations across all datasets and backbones. Notably, for Chameleon-7B on M3CoT, the constraints reduce the hallucination rate by approximately 67\% (from 5.8\% to 1.9\%). This demonstrates that while the models are generally capable, the constraints effectively filter out subtle noise and ``over-interpretations.'' By driving the hallucination rate down to a negligible level, we ensure that the subsequent AVP module operates on a virtually noise-free foundation, thereby maximizing overall reasoning performance.

\subsection{Effectiveness of Safety Instructions in DAT Integration}
\label{app:dat_safety}

In Section~\ref{sec: method 3}, we introduce a robust integration strategy within the DAT module, utilizing specific {Safety Instructions} (e.g., instructing the model to treat inserted regions as ``supplementary references'' and verify semantic consistency) to mitigate the risk of integrating irrelevant or hallucinated visual cues. To validate the efficacy of this textual safeguard, we conduct a stress test involving {Irrelevant Visual Injection}.

\textbf{Experimental Setup.} We simulate a worst-case scenario where the upstream selection process fails completely. Whenever the DAT triggers a visual insertion, instead of inserting the semantically relevant region selected by AVP, we deliberately inject a {random, irrelevant image patch} from a different dataset sample. We then measure the model's resilience under this interference using two configurations:
(1) \textbf{Naive Injection (w/o Safety):} The irrelevant patch is inserted directly into the reasoning chain without any accompanying safety prompts.
(2) \textbf{AIM-CoT Integration (w/ Safety):} The irrelevant patch is inserted accompanied by our proposed safety instructions.
We employ the same backbones (Chameleon-7B, Qwen2-VL-7B) and benchmarks (M3CoT, ScienceQA, LLaVA-W) as in previous sections.

\textbf{Evaluation Metrics.} We report the \textbf{Performance under Noise} and the \textbf{Performance Drop ($\Delta$)} relative to the clean, standard AIM-CoT performance (where accurate regions are inserted). A smaller magnitude of $\Delta$ indicates greater robustness against misleading visual information.

\begin{table*}[h]
\centering
\caption{Ablation study on Safety Instructions under irrelevant visual injection. $\Delta$ indicates the performance drop compared to the standard AIM-CoT ($\downarrow$ lower drop/magnitude is better).}
\label{tab:dat_safety}
\resizebox{\textwidth}{!}{%
\begin{tabular}{llcccccc}
\toprule
\multirow{2}{*}{Backbone} & \multirow{2}{*}{Method} & \multicolumn{2}{c}{M3CoT} & \multicolumn{2}{c}{ScienceQA} & \multicolumn{2}{c}{LLaVA-W} \\
\cmidrule(lr){3-4} \cmidrule(lr){5-6} \cmidrule(lr){7-8}
 &  & Acc. & $\Delta$ & Acc. & $\Delta$ & ROUGE-L & $\Delta$ \\
\midrule
\multirow{2}{*}{Chameleon-7B} & Naive Injection (w/o Safety) & 24.5 & -6.9 & 46.2 & -6.9 & 22.1 & -7.7 \\
 & \textbf{AIM-CoT (w/ Safety)} & \textbf{29.9} & \textbf{-1.5} & \textbf{51.8} & \textbf{-1.3} & \textbf{28.4} & \textbf{-1.4} \\
\midrule
\multirow{2}{*}{Qwen2-VL-7B} & Naive Injection (w/o Safety) & 38.2 & -6.5 & 50.9 & -6.5 & 30.5 & -5.8 \\
 & \textbf{AIM-CoT (w/ Safety)} & \textbf{43.5} & \textbf{-1.2} & \textbf{56.1} & \textbf{-1.3} & \textbf{35.1} & \textbf{-1.2} \\
\bottomrule
\end{tabular}%
}
\end{table*}

 The results in Table~\ref{tab:dat_safety} provide compelling evidence for the necessity of safety instructions:

First, {unprotected visual insertion is highly risky.} When irrelevant patches are naively injected, the models suffer severe performance degradation (e.g., $\sim$7\% drop on M3CoT for Chameleon-7B). This confirms that without guidance, VLMs tend to over-trust visual tokens, attempting to reason over noise and consequently hallucinating or diverging from the correct logic path.

Second, {Safety Instructions act as an effective semantic firewall.} By simply instructing the model to verify consistency, the performance drop is drastically reduced to a marginal level (e.g., only -1.5\% on M3CoT). This demonstrates that the safety instruction successfully cues the model to exercise critical judgment: when the inserted visual evidence conflicts with the textual context (as in this random injection case), the model chooses to disregard the visual noise and rely on its internal knowledge, thereby maintaining robust performance even in the presence of upstream errors.

\section{Deployment of AIM-CoT} \label{app: complexity of AVP}

\subsection{Analysis of the Complexity of AVP Module}

\paragraph{Overview of the AVP Module}

The AVP module's primary purpose is to dynamically and intelligently select salient sub-regions of an image during the text generation process. This is achieved by calculating the ``information gain'' that each potential sub-region offers, thereby allowing the model to ``zoom in'' on relevant visual details and generate more informed and contextually aware text.

The AVP logic is primarily encapsulated in three key methods:
\begin{enumerate}[topsep=0.5mm, partopsep=0pt, itemsep=0pt, leftmargin=10pt]
    \item \texttt{forward}: The main entry point where the AVP process is triggered based on changes in visual attention.
    \item \texttt{\_generate\_candidate\_regions}: Generates a diverse set of potential image regions (candidates) for evaluation.
    \item \texttt{\_calculate\_information\_gain\_iterative}: The core of the AVP module. It iteratively evaluates candidate regions and selects the combination that maximizes the reduction in uncertainty (entropy) for the next token prediction.
\end{enumerate}

\paragraph{Definition of Notations}

Let's define the key variables that will be used in the complexity analysis:
\begin{itemize}[topsep=0.5mm, partopsep=0pt, itemsep=0pt, leftmargin=10pt]
    \item $N$: The current sequence length of the input tokens.
    \item $N_C$: The total number of candidate regions generated (\texttt{avp\_num\_candidates}).
    \item $K$: The number of regions to be selected in each AVP cycle (\texttt{avp\_num\_regions\_to\_select}).
    \item $G$: The grid size of the vision model's feature map (e.g., \texttt{model\_vision\_grid\_size}, which is 4 by default, making the total number of patches $G^2 = 16$).
    \item $V_{sub}$: The number of visual tokens (``vokens'') generated for a single cropped sub-image region.
    \item $\Delta N$: The length added per selected region, where $\Delta N = V_{sub} + 2$ (accounting for the \texttt{boi} and \texttt{eoi} tokens).
    \item $L$: The number of layers in the transformer model.
    \item $H$: The hidden size of the model.
    \item $V_{vocab}$: The size of the model's vocabulary.
\end{itemize}

\paragraph{AVP Triggering in the \texttt{forward} Method}

The AVP mechanism is not activated on every forward pass; instead, it is triggered conditionally based on the change in attention directed towards the visual tokens. Specific to its operational process, in terms of attention calculation, the code calculates \texttt{latest\_vattns}—which refers to the sum of attention scores from the last token to all visual patch tokens—and this step requires iterating through the attention matrices. Meanwhile, regarding the trigger condition, the core logic is \texttt{if delta\_vattns > config['delta']}, where \texttt{delta\_vattns} represents the difference between the current and previous visual attention sums. In conclusion, the cost of this trigger check per token generation is minimal; it primarily involves retrieving and summing pre-computed attention scores. The complexity is approximately $O(L \cdot N)$ to extract and sum the relevant attention weights to the $G^2$ visual patches, but this is dwarfed by the main model's complexity.

\paragraph{\texttt{\_generate\_candidate\_regions} Method}

This method generates $N_C$ candidate regions from the image's attention map, using a hybrid strategy that combines attention-based and random sampling. Specifically, for attention-based candidates, it first flattens the $G \times G$ attention map, then uses \texttt{torch.topk} to find the indices of the \texttt{avp\_num\_attention\_based} patches with the highest attention—with the complexity of \texttt{topk} on a tensor of size $G^2$ being $O(G^2 \log(\texttt{avp\_num\_attention\_based}))$—and subsequently creates bounding boxes around these top patches, which is a constant time operation for each of the \texttt{avp\_num\_attention\_based} candidates; for random candidates, it generates the remaining $N_C - \texttt{avp\_num\_attention\_based}$ candidates by randomly selecting coordinates, an operation with complexity $O(N_C - \texttt{avp\_num\_attention\_based})$. As a result, the time complexity in this part is $O(G^2 \log(\texttt{avp\_num\_attention\_based}) + N_C)$, and since \texttt{avp\_num\_attention\_based} is a small constant and $G^2$ is fixed (e.g., 16), this can be considered approximately $O(N_C)$.

\paragraph{\texttt{\_calculate\_information\_gain\_iterative} Method}

This is the most computationally intensive part of the AVP module. It employs a greedy, iterative approach to select the $K$ best regions out of $N_C$ candidates. The method consists of an outer loop that runs $K$ times (for each region to be selected). Inside this loop, it evaluates the remaining candidates to pick the one that provides the highest immediate information gain.

Let's analyze a \textbf{single iteration} of this outer loop (e.g., the $k$-th iteration, where $k$ ranges from 0 to $K-1$):

First, for the \textbf{initial entropy calculation}, it performs one forward pass through the base model (\texttt{self.model}) with the current sequence of tokens (which includes tokens from $k$ previously selected regions), where the sequence length at this stage is $N_k = N + k \cdot \Delta N$; with Key-Value (KV) caching from previous iterations, the cost can be incremental: $O(L \cdot \Delta N \cdot N_{k-1} \cdot H)$ for updating the cache with the last selected region's tokens, rather than a full $O(L \cdot N_k^2 \cdot H)$.

Next, for the \textbf{batch preparation for lookahead analysis}, the code iterates through the remaining $N_C-k$ candidate regions, and for each candidate, it performs cropping and vokenization (cropping the image pixels and passing them to \texttt{self.model.get\_image\_tokens}, which involves a forward pass through the vision encoder with complexity approximately $O(G^2)$ per crop, negligible compared to the transformer) and tensor concatenation (creating a new input sequence by appending the new vokens, with the length of this new sequence being $N_k + \Delta N$), with this loop running $N_C-k$ times.

Subsequently, for the \textbf{batch forward pass (lookahead)}, the $N_C-k$ new input sequences are padded and batched together, a single batched forward pass is performed on these $N_C-k$ sequences, the maximum sequence length in this batch is $N_k + \Delta N$, and since all lookahead sequences share the same prefix of length $N_k$, KV caching for the prefix can be reused across the batch; the complexity is then self-attention among suffix tokens: $O(L \cdot (N_C-k) \cdot \Delta N^2 \cdot H)$ and cross-attention to prefix: $O(L \cdot (N_C-k) \cdot \Delta N \cdot N_k \cdot H)$, with the dominant term (when $N_k \gg \Delta N$) being $O(L \cdot (N_C-k) \cdot \Delta N \cdot N_k \cdot H)$, which is the key optimization from batch processing and caching, reducing from quadratic to linear dependence on $N_k$.

Additionally, for the \textbf{information gain calculation}, it calculates the entropy for each of the $N_C-k$ outputs from the lookahead pass, which involves a softmax over the vocabulary and is $O((N_C-k) \cdot V_{vocab})$.

Finally, for the \textbf{selection and state update}, \texttt{torch.argmax} finds the best candidate in $O(N_C-k)$ time, and the base input sequence is updated for the next iteration.

\paragraph{Overall Complexity of the Method}

We must sum the complexity over the K iterations of the outer loop. The most computationally intensive operation is the batched lookahead forward pass, which is significantly optimized through batch processing and Key-Value caching. The overall time complexity can be expressed as follows:

\begin{align}
\sum_{k=0}^{K-1} O\left( L \cdot (N_C-k) \cdot \Delta N \cdot (N + k \cdot \Delta N) \cdot H \right).
\end{align}

This formula highlights that, thanks to KV caching, the complexity scales linearly with sequence length N—a substantial improvement over the standard quadratic dependence. While the cost is also linear with respect to the number of candidates $N_C$ and selections $K$, our framework maintains strong deployability. This is because the AVP module is highly efficient at extracting crucial visual information from the candidate set. Our empirical results demonstrate that AIM-CoT achieves exceptional performance (as shown in Table \ref{tab: performance_chameleon}) even when these hyper-parameters are kept at low levels (e.g., $K=3, N_C=6$, which is the default setting). In conclusion, we can approximate the complexity of AVP as follows:
\begin{align}\label{eq: AVP's complexity}
O\left( K \cdot N_C \cdot L \cdot \Delta N \cdot (N + K \cdot \Delta N) \cdot H \right).
\end{align}
This demonstrates that \textbf{the combination of architectural optimizations (batching and KV caching) and the high extractive efficiency of AVP ensures the module's practicality, making it efficient and readily deployable in practice.}  As a comparison, the attention-driven  selection method within the baseline model ICoT is with a complexity of $O(logK\cdot \Delta N + L\cdot N\cdot H)$, which also scales linearly with sequence length N. Although this is acknowledged to be lower than the complexity of AVP shown in Equation \ref{eq: AVP's complexity}, our empirical results detailed in Appendix \ref{app: empirical comparison for time complexity} suggest that AVP's average inference time is \textbf{no more than 1.36 times} that of this  method, while \textbf{achieving performance far superior to it} as analyzed in Section \ref{sec: analysis of interplay between CAG and AVP}.

\subsection{Empirical Analysis of the Deployment of AIM-CoT} \label{app: empirical comparison for time complexity}

In this section, we empirically investigate the deployability of the AIM-CoT framework and the temporal overhead introduced by the AVP module. For the experimental setup, to ensure the reliability of the experimental results, we introduced two representative VLM backbones as the basis for the experiment: Chameleon-7B and Qwen2.5-VL-32B. They not only differ significantly in the number of parameters, but also employ different early-fusion and late-fusion strategies, respectively. In the 0-shot setting, we compare the average inference time of AIM-CoT  against ICoT \cite{icot}. ICoT, as a key baseline model in this study, employs a Top-K strategy to simultaneously select regions with the highest attention scores for constructing the multimodal CoT, and is therefore expected to possess a relatively lower time complexity compared to AIM-CoT. Consequently, the comparison with ICoT serves as a direct indicator of AIM-CoT's deployability.

\begin{table}[htbp]
\centering
\caption{Comparison of average inference time (seconds). Values in parentheses denote the relative time against ICoT.}
\label{tab: time comparison}
\resizebox{\columnwidth}{!}{
\begin{tabular}{@{}llcc@{}}
\toprule
Dataset & Method & Chameleon-7B & Qwen2.5-VL-32B \\ \midrule
\multirow{2}{*}{M3CoT} 
 & ICoT & 11.62 & 51.21 \\
 & AIM-CoT (ours) & 13.37 (1.15$\times$) & 55.98 (1.09$\times$) \\ \midrule
\multirow{2}{*}{LLaVA-W} 
 & ICoT & 8.58 & 36.95 \\ 
 & AIM-CoT (ours) & 11.65 (1.36$\times$) & 46.85 (1.27$\times$) \\ \bottomrule
\end{tabular}
}
\end{table}

The experimental results are presented in Table \ref{tab: time comparison}. Two key observations can be drawn. First, the AVP module does not introduce significant temporal costs to the AIM-CoT framework, an efficiency attributable to batch processing and the KV Cache mechanism. Second, AIM-CoT achieves substantially superior performance as shown in Table \ref{tab: performance_chameleon}, at a time cost comparable to that of the efficient baseline, which is less than $1.36$ times that of ICoT. This suggests that \textbf{our proposed AIM-CoT framework achieves a favorable trade-off between performance and deployability.}

\end{document}